\documentclass[letterpaper]{article} 
\usepackage{aaai23}  
\usepackage{times}  
\usepackage{helvet}  
\usepackage{courier}  
\usepackage[hyphens]{url}  
\usepackage{graphicx} 
\usepackage{natbib}  
\usepackage{caption} 
\usepackage{algorithm}
\usepackage{algorithmic}

\usepackage{newfloat}
\usepackage{listings}

\usepackage{amsmath}
\usepackage{graphicx}
\usepackage{multirow}
\usepackage{soul}

\usepackage{pifont}

\usepackage{times}
\usepackage{latexsym}
\usepackage{graphicx}
\usepackage{booktabs}
\usepackage{subcaption}
\usepackage{multirow}
\usepackage{tabularx}
\usepackage{amsmath}
\usepackage{bbold}
\usepackage{mathtools}
\usepackage{xcolor}
\usepackage{booktabs}
\usepackage{tabularx,ragged2e}
\usepackage{enumitem}

\DeclareCaptionStyle{ruled}{labelfont=normalfont,labelsep=colon,strut=off} 
\floatstyle{ruled}
\newfloat{listing}{tb}{lst}{}
\floatname{listing}{Listing}
\title{Can We Identify How much external knowledge is sufficient to answer open-domain natural questions? Investigating its Positive Impact on System's Efficiency }

\title{[Temp] Investigating Use of `On Need Basis' External Knowledge to Answer Open-Domain Natural Questions and its Impact on Reader Efficiency}

\title{Investigating Efficient Use of External Knowledge in Answering Open-Domain Natural Questions and its Impact on Reader Efficiency}

\title{Can Open-Domain QA Reader Efficiently Utilize the External Knowledge without Sacrificing the Prediction Performance?}

\title{Can Open-Domain QA Reader Make Efficient Use of External Knowledge?}

\title{Can Open-Domain QA Reader Make Efficient Use of External Knowledge like Humans?}

\title{Can Open-Domain QA Reader Utilize External Knowledge Efficiently like Humans?}

\author {
    Neeraj Varshney, Man Luo, Chitta Baral
}
\affiliations{
    Arizona State University \\
    \{nvarshn2, mluo26, chitta\}@asu.edu
}
\usepackage{bibentry}

\begin{document}

\maketitle

\begin{abstract}
Recent state-of-the-art open-domain QA models are typically based on a two stage retriever–reader approach in which the retriever first finds the relevant knowledge/passages and the reader then leverages that to predict the answer.
Prior work has shown that the performance of the reader usually tends to improve with the increase in the number of these passages. 
Thus, state-of-the-art models use a large number of passages (e.g. 100) for inference.
While the reader in this approach achieves high prediction performance, its inference is computationally very expensive.
We humans, on the other hand, use a more efficient strategy while answering: 
firstly, if we can confidently answer the question using our already acquired knowledge then we do not even use the external knowledge, and in the case when we do require external knowledge, we don't read the entire knowledge at once, instead, we only read that much knowledge that is sufficient to find the answer.
Motivated by this procedure, we ask a research question ``\textit{Can the open-domain QA reader utilize external knowledge efficiently like humans without sacrificing the prediction performance?}''

Driven by this question, we explore an approach that utilizes both the `closed-book' (leveraging knowledge already present in the model parameters) and the `open-book' inference (leveraging external knowledge).
Furthermore, instead of using a large fixed number of passages for open-book inference, we dynamically read the external knowledge in multiple `knowledge iterations'.
Through comprehensive experiments on NQ and TriviaQA datasets, we demonstrate that this dynamic reading approach improves both the \textbf{inference efficiency} and the \textbf{prediction accuracy} of the reader.
Comparing with the top-performing Fusion-in-Decoder reader, this approach matches its accuracy by utilizing just $18.32\%$ of its reader inference cost (measured in FLOPs) and also outperforms it by achieving up to $55.10\%$ accuracy on NQ Open.

\end{abstract}

\section{Introduction}

Recently developed retriever-reader systems \cite{chen-etal-2017-reading, karpukhin-etal-2020-dense, 10.1145/3397271.3401075,izacard-grave-2021-leveraging} have achieved impressive performance on the open-domain question answering task. 
In this pipeline, the retriever finds the top-$N$ relevant passages and the reader model leverages them to predict the answer. 
Prior work has shown that the performance of the reader tends to improve (up to a certain extent) with the increase in the value of $N$.
Thus, state-of-the-art models use a large number of passages (e.g. 100).
While this strategy results in a high prediction performance, it makes the inference of the reader computationally very expensive.
For instance, Fusion-in-Decoder reader model (FiD) \cite{izacard-grave-2021-leveraging} requires approximately $70$ $\times$ $10^{11}$ floating-point operations (FLOPs) for an inference with $100$ passages.
This high inference cost limits the widespread adoption of such systems in real-world applications that prefer efficient systems to be able to achieve low response times.


Improving the efficiency of systems has been an important research topic in NLP.
For the open-domain QA (ODQA) task, efficiency from the perspectives of retrieval \cite{zhao-etal-2021-sparta} and on-disk memory \cite{pmlr-v133-min21a, izacard2020memory,yamada-etal-2021-efficient} has been studied.
However, the aspect of efficiently leveraging external knowledge to improve the computation performance of the reader model has remained underexplored.


Motivated by the efficient approach that humans typically use for answering questions, we argue that the reader model does not always require all the top-$N$ passages to answer a question correctly. 
Some questions are trivial and can be answered with even a few passages or even without using any external knowledge at all (by just relying on the knowledge already stored in the model parameters).
Consider the case of FiD model, it requires $\sim70$ $\times$ $10^{11}$ FLOPs for an inference with $100$ passages and $\sim7$ $\times$ $10^{11}$ FLOPs for an inference with $10$ passages.
On the Natural Questions dataset (NQ), it achieves $54.43\%$ and $50.61\%$ exact match accuracies in the former and the latter cases respectively.
From this, it is clear that the performance of the system utilizing $100$ passages can be matched by inferring a large number of instances with just $10$ passages and only a few instances with $100$ passages.
Furthermore, the closed-book model \cite{roberts-etal-2020-much} that does not use any external knowledge and relies only on the knowledge in its parameters (acquired during pre-training/fine-tuning) requires just $\sim0.06$ $\times$ $10^{11}$ FLOPs and achieves lower yet non-trivial accuracy of $29.83\%$. 
\textbf{Thus, by carefully deciding when external knowledge is required and whether the current amount of external knowledge is sufficient to answer a question correctly, the computational efficiency of the reader system can be considerably improved while maintaining the high prediction accuracy.}
Moreover, this can also help the system mitigate the distraction that may result from using too many passages for inference and thus can even improve its prediction accuracy.






Following the above intuition, we explore an approach that utilizes both `closed-book' and `open-book' inferences and dynamically uses external knowledge in multiple \textit{knowledge iterations}.
Specifically, given a question, we first infer it using the low-cost closed-book model that relies on the knowledge already stored in its parameters. 
If its prediction confidence is sufficiently high then the prediction is outputted otherwise we infer using the open-book model leveraging the external knowledge. 
Unlike the standard open-book models that always `read' a fixed number of passages, in our method, the knowledge provided is iteratively increased until its prediction becomes sufficiently confident.
We demonstrate that the confidence of prediction shows a positive correlation with the predictive correctness. 
Thus, instances that are predicted with high confidence using low-cost inference get answered at early stages as their predictions are likely to be correct, and the remaining instances get answered with dynamically used external knowledge.
Hence, by avoiding expensive inference primarily for easy instances and dynamically using external knowledge, our approach makes the reader system computationally efficient while maintaining high prediction accuracy.


Through comprehensive experiments on NQ and TriviaQA datasets, we first show that our approach considerably improves the computational efficiency of inference of the reader model. 
Comparing with a top-performing FiD reader, we show that our approach matches its accuracy by utilizing just $18.32\%$ of its inference computation cost (measured in FLOPs).
Then, we show that our approach also leads to a consistent improvement in prediction accuracy.
Specifically, it outperforms FiD by achieving up to $55.10\%$ accuracy on NQ Open.
This improvement is an outcome of mitigating distraction that results from using too many passages for inference.
Finally, we note that our approach is intuitive, easy to implement, and also has practical values.

\section{Related Work}

In recent times, a considerable research effort has been invested in improving the efficiency of NLP systems. It has lead to development of several techniques, such as 
\textit{network pruning} \cite{wang-etal-2020-structured,guo-etal-2021-parameter}, 
\textit{quantization} \cite{shen2020q,zhang-etal-2020-ternarybert,tao2022compression},
\textit{knowledge distillation} \cite{clark-etal-2019-bam,jiao-etal-2020-tinybert,li2022dq,mirzadeh2020improved}, dynamic inference \cite{xin-etal-2020-deebert}, adaptive model size \cite{goyal2020power, kim-cho-2021-length, NEURIPS2020_6f5216f8}, cascading \cite{xin-etal-2021-art,varshney2022model}, and
\textit{input reduction} \cite{modarressi2022adapler}.



In open-domain QA, efficiency from the perspectives of retrieval \cite{zhao-etal-2021-sparta,luo2022study} and on-disk memory \cite{pmlr-v133-min21a, izacard2020memory} has been explored.
Iterative \cite{das2018multistep,qi-etal-2019-answering} and adaptive \cite{kratzwald-feuerriegel-2018-adaptive} retrieval are other related research fields that aim to search for the relevant documents from a large collection in multiple steps.

We study an important yet underexplored area of efficiently leveraging external knowledge to improve the computation efficiency of inference of the reader system.
Our work differs from existing work in the following aspects:
(1) Firstly, the inference efficiency aspect has mostly been studied for the classification tasks using encoder-based models. However, we focus on a more challenging task of open-domain QA with generative models. 
To the best of our knowledge, leveraging the prediction confidence in generative models for improving the efficiency of inference has not been explored previously. 
(2) Existing efficiency improvement methods typically require architectural changes, network manipulation, saliency quantification, knowledge distillation, or even complex training procedures. In contrast, our method is easy to implement, does not require such modifications, and could generalize easily to a variety of applications. Moreover, it can even complement these existing methods.
(3) The computation efficiency in existing methods often comes with a compromise on accuracy. In contrast, we show that our method consistently achieves superior accuracy.
(4) Finally, existing methods usually do not allow custom computation costs and require training a separate model for each computation budget. In contrast, our system can be adjusted to meet the given computational requirements.

\section{Approach}

In open-domain QA, the cost of reader inference depends on the number of passages used as additional context for the inference. 
Prior work has shown that the performance of the reader tends to improve (up to a certain extent) with the increase in the number of these passages.
Thus, state-of-the-art models use a large number of passages (e.g. 100).
While this strategy results in a high prediction performance, it makes the inference of the reader computationally very expensive.
Motivated by the strategy that humans typically use for answering questions, we explore an approach that efficiently leverages the external knowledge by dynamically using it in multiple \textit{knowledge iterations}.

Firstly, we note that even a \textbf{closed-book} reader (CB) that does not use any external knowledge achieves a non-trivial accuracy by just relying on the knowledge already stored in its network parameters (acquired during pre-training/fine-tuning).
This is a low-cost inference as the input contains just the question (without any additional context).
So, in our approach, we first infer the given question using the closed-book reader and output the prediction if it is already sufficiently confident. 
If it is not confident then we leverage the external knowledge with the \textbf{open-book} reader (OB).
Unlike the standard open-book readers that use all the top retrieved passages for inference, we iteratively increase the number of passages until the reader predicts with sufficient confidence; we refer to these iterations as `\textbf{knowledge iterations}'.

This conditional multi-stage inference process achieves computation efficiency benefits for two reasons: first, if CB reader is already sufficiently confident in its prediction then the expensive open-book inference is not used at all (this corresponds to the case of the least inference cost) and second, when OB reader is indeed used for inference, it leverages the external knowledge efficiently by reading just enough passages required to predict confidently instead of reading a static large number of passages.
This approach can also help the system mitigate the distraction that may result from using too many passages for inference.

Hence, by avoiding expensive inference and dynamically using the external knowledge, our approach makes the reader inference computationally efficient while maintaining high prediction accuracy.
We note that this \textbf{doesn't impact the retriever as the retrieval is done only once} irrespective of the number of knowledge iterations and the number of passages used in each individual iteration.
We further note that \textbf{in this work, our focus is only on improving the cost of reader inference}.
We detail our approach and provide its mathematical formulation in the next subsection.




\subsection{Mathematical Formulation}
\label{subsec_formulation}


Let $q$ be the given question, $K$ be the number of knowledge iterations, and $M_{{OB}_k}^{q}$ be the value indicating whether $k^{th}$ iteration ($k \leq K$) with the OB reader is used for inference.

\begin{equation*}
        M_{{OB}_k}^{q} =
            \begin{cases}
              1, & \text{if $k^{th}$ knowledge iteration with } \\
              & \text{OB reader is used for question $q$} \\
              0, & \text{otherwise}
              
            \end{cases}
\end{equation*}
\paragraph{Sample Scenario:}
Consider a scenario in which top-100 relevant passages are available and in our approach we are using two knowledge iterations ($K=2$) using $20$ and $100$ passages in the two iterations respectively i.e. a question will be first inferred using the CB reader, if it is not sufficiently confident then top-20 (referred as $S_1$) passages will be used with the OB model (referred as $OB_1$) and if that prediction is not confident then top-100 ($S_2$) passages will be used with the OB model ($OB_2$).
In the computationally most efficient case, inference would be made only using the CB reader and in the most expensive case, inference would be made sequentially using CB, then $OB_1$ (OB using 20 passages), and then $OB_2$ (OB using 100 passages). In this work, we comprehensively study extensive combinations of CB, OB readers and values of $K$ and $S_k$.

\paragraph{Cost of Reader Inference: } 
In our general formulation, the cost of reader inference for instance $q$ is calculated as:
\[
      Cost^{q} = C_{CB} + \sum_{k=1}^{K} (M_{{OB}_k}^{q} \times S_k \times C_{OB})
\]
where $C_{CB}$ and $C_{OB}$ are the inference costs of CB and OB models respectively, and $S_k$ corresponds to the number of passages used by the OB model in $k^{th}$ knowledge iteration.
This is because the low-cost CB reader is first used for all the instances and then OB model is conditionally used in different knowledge iterations.
In the sample scenario, the minimum cost would be equal to $C_{CB}$ and the maximum cost would be $C_{CB} + 20 \times C_{OB} + 100 \times C_{OB}$.
Next, we discuss a few important characteristics of this formulation.


\paragraph{Inference Cost of OB Reader:}
Fusion-in-Decoder model \cite{izacard-grave-2021-leveraging} is one of the top performing open-book models. It uses an encoder-decoder architecture i.e. it first computes the representation of question + passage for each passage independently using the encoder (with fixed number of input tokens) and then concatenates these representations and passes it to the decoder for making the answer prediction.
Thus, to compute the OB model's cost of inference in the $k^{th}$ iteration, we multiply $C_{OB}$ with $S_k$ where $C_{OB}$ is the cost of single inference with that fixed number of tokens and it is done $S_k$ number of times. 

However, we note that this is the upper bound of the inference cost because the encoder representation of the passages of the previous iterations can be reused i.e. the representations of passages of $(k-1)^{th}$ iteration can be reused in the $k^{th}$ iteration.
However, this would require auxiliary space for storage and would involve a trade-off between computation cost and storage space. We leave the investigation of this trade-off for future work but \textbf{note that the inference cost of our method would be even lower in practice}.
Moreover, as we demonstrate through extensive experiments that even with this cost upper bound, the proposed method achieves very high improvements in computation efficiency.



\paragraph{Same Model for CB and OB:}
We note that the same model can also be used to act as CB when external knowledge is not available and as OB when it is available. 
However, their cost of inference will still be different as it depends on the number of input tokens used for inference.
The CB reader uses only the question as input while the OB reader also uses the external knowledge (thus more number of input tokens).
Therefore, to keep our formulation general, we keep two different variables for their respective costs.



\paragraph{Total Cost of Reader Inference: }
We calculate the average cost of reader inference for the evaluation dataset $D$ as:
\[
     Cost_{D} = \frac{\sum_{q_j \in D} Cost^{q_j}}
     {|D|} 
\]






\subsection{Deciding When to Use More Knowledge}

\label{subsec_computing_confidence}

The leading models in ODQA and perhaps in many NLU tasks are nowadays mostly seq2seq generative models; for e.g. both the closed-book and FiD models are based on T5 \cite{JMLR:v21:20-074}. 
Therefore, the main design decision in our method corresponds to computing the confidence of prediction for these models. 
These models make their predictions token by token and output a probability distribution over the entire vocabulary for each token.
We explore a number of methods to compute the confidence scores.


Let the maximum softmax probabilities for the prediction having $n$ tokens at each token position be $(p_1, p_2, p_3, ..., p_n)$.
We explore the following ways of computing the model's prediction confidence using these $p$ values:

\paragraph{Product of probabilities of all tokens ($P_{PA}$):}
In this technique, we take the product of probabilities of all tokens of the prediction i.e. $PROD(p_1, p_2, ..., p_n)$ as the model's prediction confidence.
This is the standard technique used in various tasks such as perplexity computation.
We also experiment with several other confidence techniques.

\paragraph{Probability of the first token ($P_F$):} In this technique, we simply use the probability of the first token of the prediction i.e. $p_1$ as the model's prediction confidence.

\paragraph{Average probability of first and last token ($P_{FL}$):}
Here, we use the average probability of the first and the last token i.e. $AVG(p_1, p_n)$ as the model's prediction confidence. 

\paragraph{Average probability across all tokens ($P_{A}$):}
In this technique, we utilize probabilities of all the tokens of the prediction and take the average probability across all tokens i.e. $AVG(p_1, p_2, ..., p_n)$ as model's prediction confidence.

\subsection{Baseline Approaches}
For a fair comparison of our approach (and various confidence computation methods), we also compare its performance with several other simple baselines:

\textbf{Random:} In this method, instead of using a metric based on probabilities to decide which instances to pass to the OB model/next knowledge iteration, we do this instance selection process at random.

\textbf{Heuristic:} Here, we use a heuristic derived from the input question to decide which instances to pass to the OB model/next knowledge iteration. Specifically, we use length of the input question as the heuristic.

\subsection{Performance Comparison Metric}
\label{subsec_perf_eval}

We demonstrate the efficacy of our method by showing the \textbf{computation efficiency} and \textbf{accuracy} improvements.
For demonstrating computation efficiency, we use FLOPs as the metric.
An alternative metric could be measuring the \textbf{computation time} of inference; however, it is a machine-dependent metric. FLOPs on the other hand, is machine-independent and hence a reliable metric for comparison.

\begin{figure*}[t]
\centering

    \begin{subfigure}{.26\linewidth}
        \includegraphics[width=\linewidth]{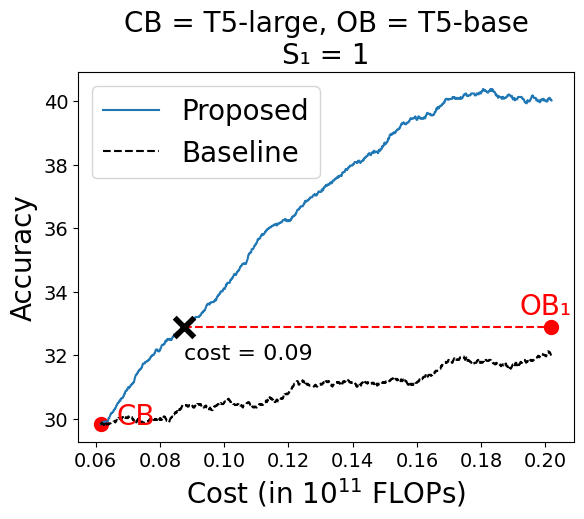}
    \end{subfigure}
    \begin{subfigure}{.28\linewidth}
         \includegraphics[width=\linewidth]{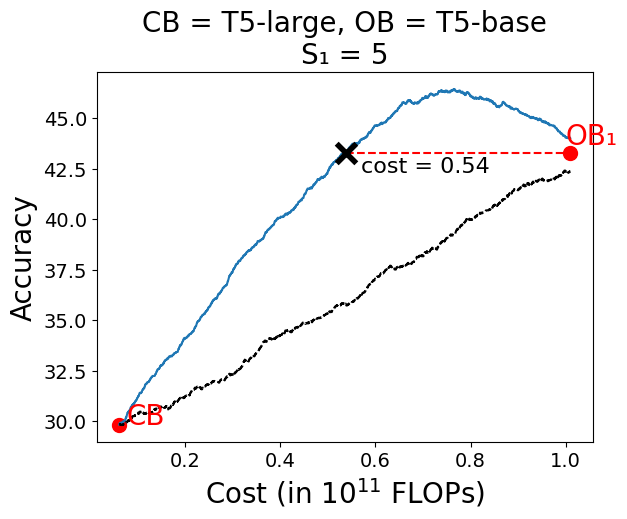}
    \end{subfigure}
    \begin{subfigure}{.28\linewidth}
         \includegraphics[width=\linewidth]{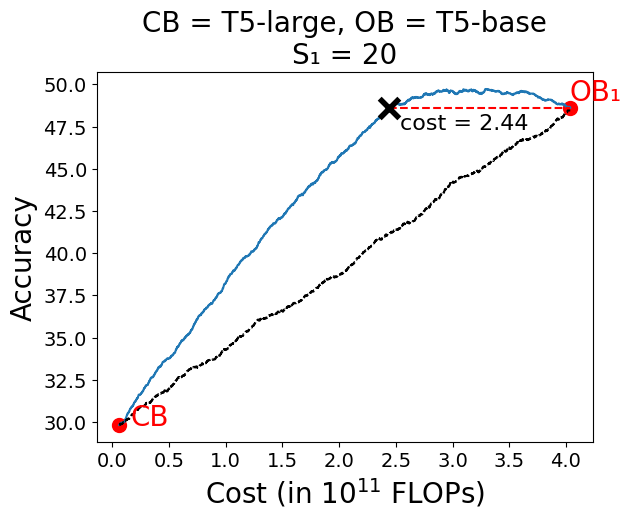}
    \end{subfigure}
    \begin{subfigure}{.27\linewidth}
         \includegraphics[width=\linewidth]{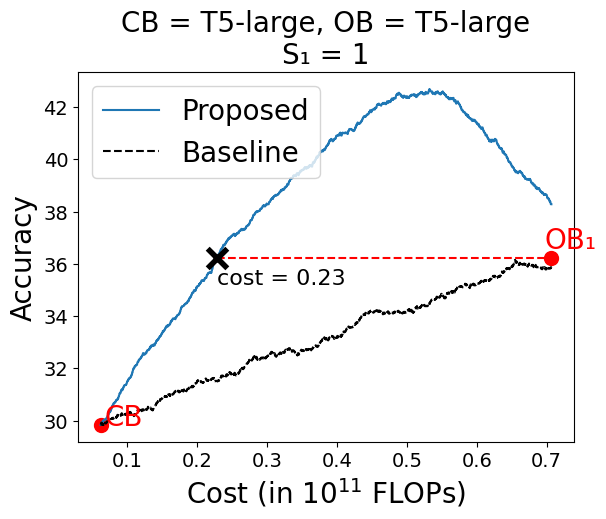}
    \end{subfigure}
    \begin{subfigure}{.28\linewidth}
         \includegraphics[width=\linewidth]{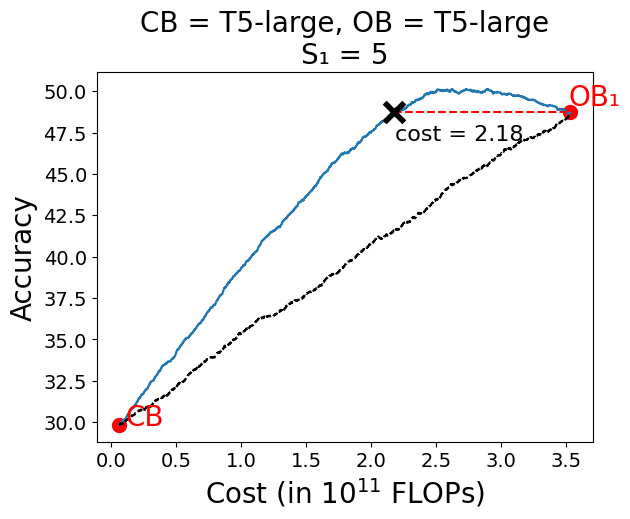}
    \end{subfigure}
    \begin{subfigure}{.28\linewidth}
         \includegraphics[width=\linewidth]{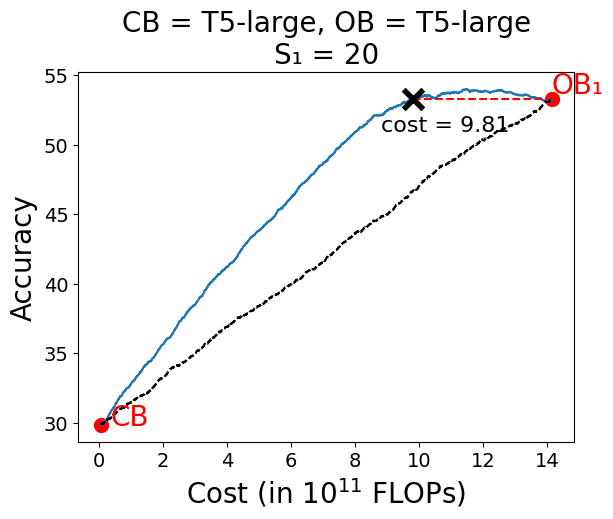}
    \end{subfigure}
    \caption{Accuracy-cost curves of the proposed system (in blue) and baseline (in black) for K=1 setting on NQ. 
    Red points correspond to the accuracy and cost values of the individual models $CB$ and $OB_1$ (using $S_1$ knowledge statements). 
    Point of intersection ($\times$) of red dashed line drawn from $OB_1$ on the blue curve corresponds to cost at which the proposed system achieves the same accuracy as $OB_1$. \textbf{Our method achieves this accuracy at considerably lower computation cost}.
    }
    \label{fig:acc_cost_curves_KI_1}    
\end{figure*}

To further compare various confidence computation methods and baselines, we use another performance metric.
For our approach, the reader inference cost (in FLOPs) and the accuracy vary with the prediction confidences thresholds.
Therefore, to compare various confidence methods and comprehensively study their efficacy, we compute accuracies for a range of costs and plot an accuracy-cost curve (as shown in Figure \ref{fig:acc_cost_curves_KI_1}). 
We plot a curve for each method and calculate the \textbf{area under the curve (AUC)} to quantify the overall performance of each method.
The larger the AUC value, the better the method is as it implies higher accuracy on average across all computation costs. 





\begin{figure*}[t]
\centering

    \begin{subfigure}{.27\linewidth}
        \includegraphics[width=\linewidth]{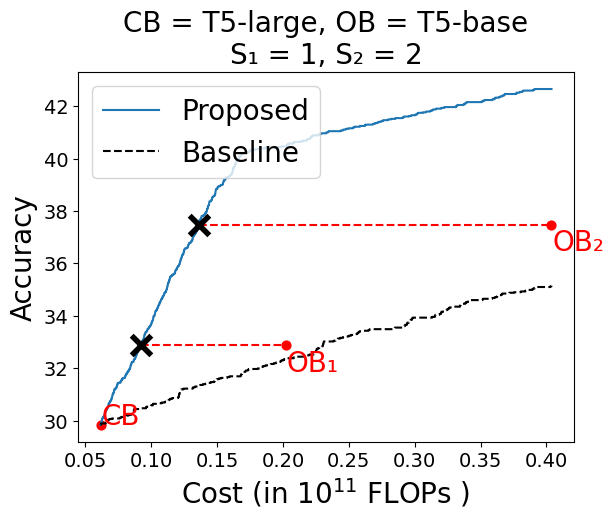}
    \end{subfigure}
    \begin{subfigure}{.28\linewidth}
         \includegraphics[width=\linewidth]{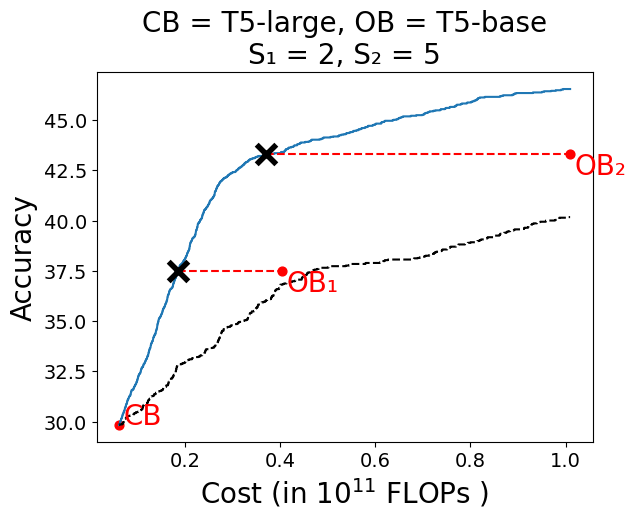}
    \end{subfigure}
    \begin{subfigure}{.3\linewidth}
         \includegraphics[width=\linewidth]{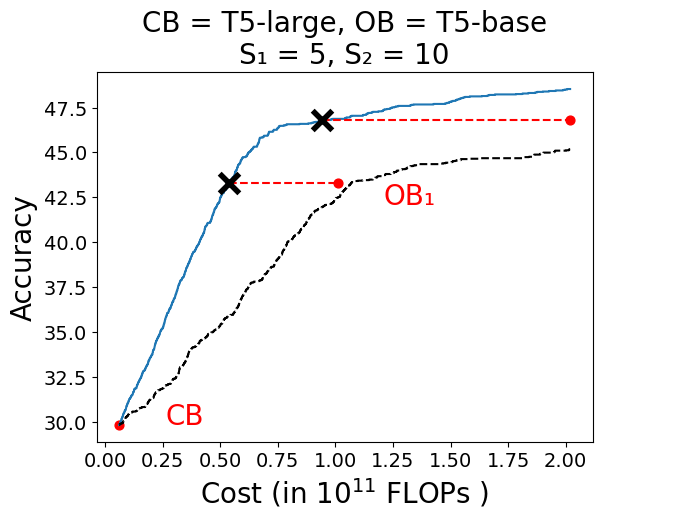}
    \end{subfigure}
    \begin{subfigure}{.28\linewidth}
         \includegraphics[width=\linewidth]{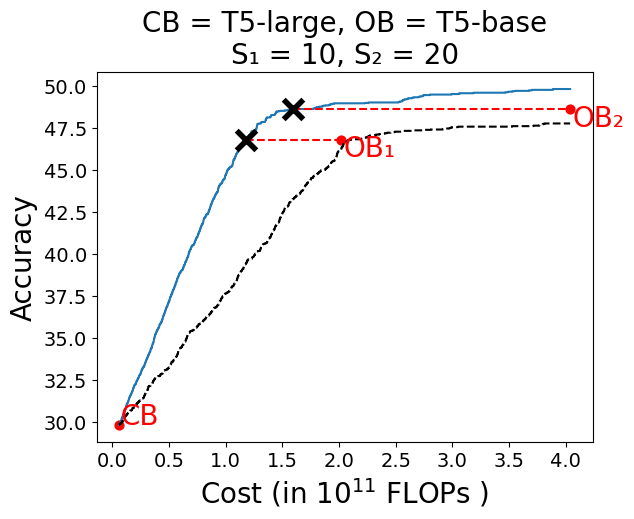}
    \end{subfigure}
    \begin{subfigure}{.28\linewidth}
         \includegraphics[width=\linewidth]{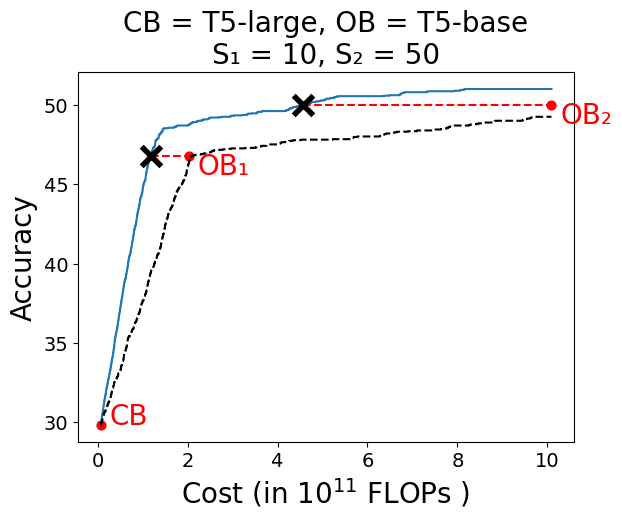}
    \end{subfigure}
    \begin{subfigure}{.28\linewidth}
         \includegraphics[width=\linewidth]{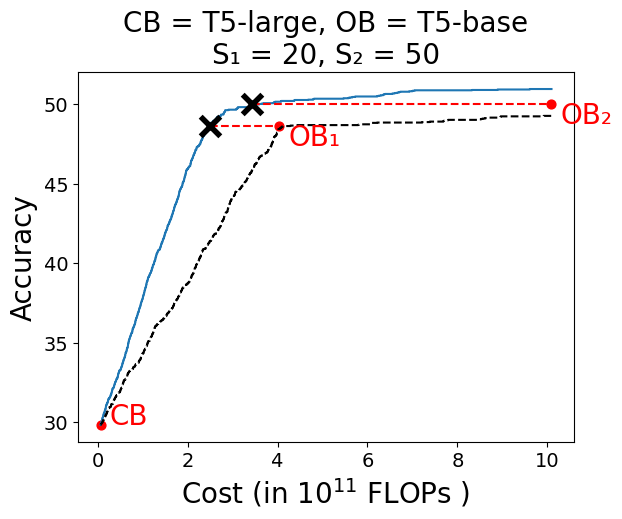}
    \end{subfigure}
    \caption{
    Accuracy-cost curves of the proposed method (in blue) and baseline (in black) for K=2 setting on NQ. 
    Red points correspond to the accuracy and cost values of the individual models $CB$, $OB_1$ (using $S_1$ knowledge), and $OB_2$ (using $S_2$ knowledge). 
    Points of intersection of red dashed lines drawn from $OB_1$ and $OB_2$ on the blue curve correspond to costs at which our system achieves the same accuracy as $OB_1$ and $OB_2$ respectively.
    }
    \label{fig:acc_cost_curves_KI_2}    
\end{figure*}

\section{Experiments and Results}
\label{sec_experiments}
\paragraph{Experimental Details: }
We conduct extensive experiments using NQ \cite{kwiatkowski2019natural} Open and TriviaQA \cite{joshi-etal-2017-triviaqa} datasets.
We use closed-book models from \citet{roberts-etal-2020-much} and FiD open-book models (and retrieved passages) from \citet{izacard-grave-2021-leveraging}.
Recall that the closed-book models take just the question as input.
The inference cost of closed-book model is $0.0046 \times 10^{11}$ FLOPs for T5-small and $0.0615 \times 10^{11}$ FLOPs for T5-large for the average input size ($12$ tokens) of NQ questions. 
On the other hand, the open-book reader also takes the retrieved passages as input (truncated to $250$ tokens each) and thus has higher inference cost.
For example, the cost of inference for the open-book FiD reader with $1$ passage is $0.202 \times 10^{11}$ FLOPs for T5-base and $0.707 \times 10^{11}$ FLOPs for T5-large. 
We compute these and other FLOP values using `thop' python library and provide further details in the Appendix.

\begin{figure*}[t]
\centering

    \begin{subfigure}{.28\linewidth}
        \includegraphics[width=\linewidth]{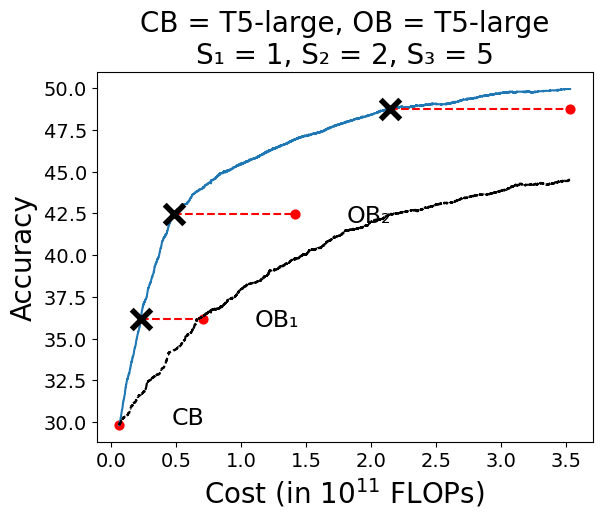}
    \end{subfigure}
    \begin{subfigure}{.28\linewidth}
         \includegraphics[width=\linewidth]{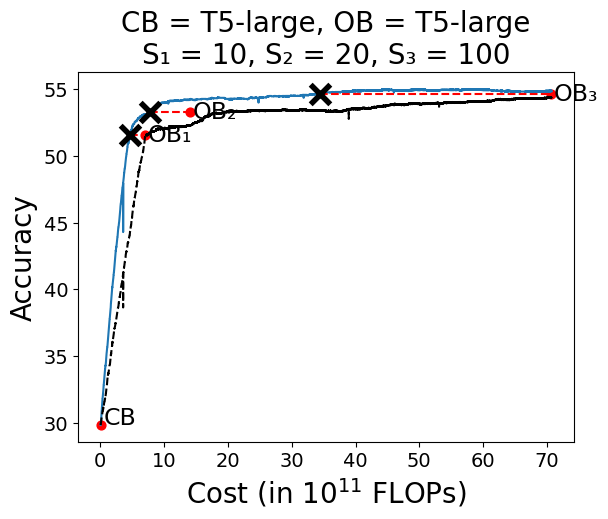}
    \end{subfigure}
    \begin{subfigure}{.28\linewidth}
         \includegraphics[width=\linewidth]{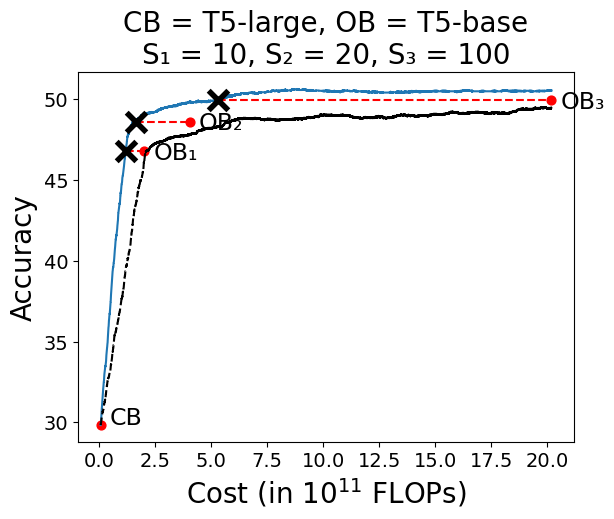}
    \end{subfigure}
    
    \caption{
    Accuracy-cost curves of the proposed method (in blue) and baseline (in black) for K=3 setting on NQ. Red points correspond to the accuracy and cost values of the individual $CB$, $OB_1$, $OB_2$, and $OB_3$ models. 
    }
    \label{fig:acc_cost_curves_KI_3}    
\end{figure*}

\paragraph{Accuracy-Cost Curves: }
As motivated previously, we plot accuracy-cost curves to study our method.
We conduct experiments in multiple settings that differ in the (CB, OB) model combination, number of knowledge iterations with the OB model ($K$), and the $S_k$ values.
Figure \ref{fig:acc_cost_curves_KI_1}, \ref{fig:acc_cost_curves_KI_2}, and \ref{fig:acc_cost_curves_KI_3} show these curves for $K$ = 1, 2, and 3 settings respectively.
Each curve includes the following: 

\begin{enumerate}[noitemsep,nosep,leftmargin=*]
    \item \textbf{Accuracy-cost points of individual systems}: 
    The costs-accuracy point of the CB and $OB_k$ models are represented by the red scatter points. These individual points correspond to the case in which all the instances are answered using the corresponding reader model.
    
    \item \textbf{Accuracy-cost curve of the proposed method}: 
    We explore multiple ways of computing prediction confidence of the generative models. 
    We compare the AUC (of their corresponding accuracy-cost curves) achieved by each method in Table \ref{tab:auc_KI_1}.
    We present an exhaustive comparison of these methods in Tables \ref{tab:auc_full_K_1_base} and \ref{tab:auc_full_K_1_large} in Appendix.
    Since \textbf{\boldmath{$P_{PA}$} yields the best performance}, we present accuracy-cost curves of the $P_{PA}$ method in the main results, and the other methods in the Appendix.
    We note that the other confidence methods also outperform the baselines. 
    
    
    \item \textbf{Costs at Equal Accuracies}: To measure the improvement in efficiency, we highlight (with {$\times$}) the costs at which the accuracy of the proposed system matches the accuracy of the $OB_k$ models.
    For instance, in K=1 setting, we highlight the point at which the proposed system achieves the same accuracy as the $OB_1$ model (that use a fixed $S_1$ number of passages for all questions).
    
    \item \textbf{Accuracy-cost curve of the baseline method}: To demonstrate the efficacy of $P_{PA}$ method in comparison to the baseline method, we represent the accuracy-cost curve of the baseline with black dashed line. \\
\end{enumerate}
In the next three subsections, we show the results for configurations with different number of knowledge iterations.


\subsection{One Knowledge Iteration (K = 1)}

In this setting, we first use the CB model and if it is not sufficiently confident then we use the OB model with $S_1$ knowledge statements.
Figure \ref{fig:acc_cost_curves_KI_1} shows the accuracy-cost curves for different configurations of CB, OB, and $S_1$ values on NQ.

\paragraph{Improvement in Efficiency:}
The accuracy-cost curves show that the proposed system matches the accuracy of the OB model at a considerably lesser inference cost. 
This cost value corresponds to the point of intersection ($\times$) on the curve with a straight horizontal line drawn from $OB_1$.
For example, in the case of (CB= T5-large, OB=T5-base, $S_1=5$), $OB_1$ achieves $43.30\%$ accuracy at the computation cost of $1.01 \times 10^{11}$ FLOPs while the proposed system achieves the same accuracy at the cost of just $0.54 \times 10^{11}$ FLOPs. 
Such improvements are observed for all the cases.
We show these curves for an exhaustive set of configurations in Appendix.
In this setting, our approach achieves cost improvements of up to $67.77\%$.
This efficiency benefit comes from using the low-cost closed-book model for some instances where it is likely to be correct and additionally using the more expensive open-book model with $S_1$ passages only for the remaining instances.
In the later results sections, we further discuss the overall improvement in the cost of reader inference (Table \ref{tab:efficiency_improvement}).

\paragraph{Improvement in Accuracy:}
From the accuracy-cost curves, it is clear that the accuracy achieved by the proposed system surpasses the accuracy of the $OB_1$ model beyond the cost shown with $\boldmath{\times}$. 
For example, in case of (CB= T5-large, OB=T5-base, $S_1=1$, the top-left figure), our system achieves the accuracy of up to $40.17\%$ as compared to $32.88\%$ accuracy of $OB_1$.
Same as the efficiency improvement, such accuracy improvements are also observed across all configurations (shown in Appendix).
We attribute this improvement in accuracy to our approach's efficient use of external knowledge i.e. relying on the closed-book model when it is likely to be correct thus avoiding distraction with the external knowledge and using it only when it is required. 

\paragraph{Distraction At Inference:}
The fact that our approach (that outputs its predictions using CB for some instances and OB for the others) outperforms the OB reader highlights that excessive external knowledge distracts the reader into giving incorrect answers while the CB reader answers it correctly.

We note that near the $OB_1$ computation cost, the accuracy of our system begins to come closer to the $OB_1$'s accuracy as more and more instances get answered by the $OB_1$ model.

\begin{table}[t]
    \centering
    \small
    \begin{tabular}{@{}l|cccccc@{}}
        \toprule
         \boldmath{$S_1 \rightarrow$} & \boldmath{$1$} & \boldmath{$2$} & \boldmath{$5$} & \boldmath{$10$} & \boldmath{$20$} & \boldmath{$50$}\\
        \midrule
        
        \textbf{Random} & 30.96 & 33.28 & 36.05 & 38.12 & 39.27 & 39.96\\
        \textbf{Heuristic} & 31.90 & 33.97 & 36.90 & 38.77 & 39.69 & 40.22\\
        \midrule
        \boldmath{$P_{PA}$} & \textbf{36.61} & \textbf{38.53} & \textbf{41.14} & \textbf{42.82} & \textbf{43.69} & \textbf{44.43}\\
        
        \boldmath{$P_F$} & 36.15 & 38.09  & 40.77 & 42.47 & 43.34 & 44.13\\
        \boldmath{$P_{FL}$} & 36.13 & 38.08 & 40.76 & 42.46 & 43.35 & 44.16\\
        \boldmath{$P_A$} & 36.39 & 38.35  & {41.00} & {42.74} & {43.56} & {44.30}\\


    \bottomrule
    \end{tabular}
    \caption{
    Comparing AUCs of accuracy-cost curves of different confidence computation techniques for CB=T5-large, OB=T5-base configuration in K=1 setting on NQ. 
    }
    \label{tab:auc_KI_1}
\end{table}
\paragraph{Comparison with Baselines:}
From the accuracy-cost curves, it is clear that our proposed method that uses $P_{PA}$ as the prediction confidence (blue curve) outperforms the baseline (black curve) as the blue curve is consistently above the black curve and thus has higher AUC value.
In Table \ref{tab:auc_KI_1}, we compare AUC values achieved by all the confidence measures: $P_F$, $P_{FL}$, $P_A$, and $P_{PA}$. 
All the approaches achieve considerably higher AUCs than the baselines while $P_{PA}$ achieves the highest.
This highlights the effectiveness of our confidence computation methods.
We show this comparison for other configurations in Appendix.
As $P_{PA}$ achieves the highest AUC, we use it for our subsequent experiments.

\begin{table}[t]
    \centering
    \small
    \begin{tabular}{@{}lcc@{}}
        \toprule
         \textbf{Method} & \textbf{NQ} & \textbf{TQA}\\
         
        \midrule
        
        Hard EM \cite{min-etal-2019-discrete} & 28.8 & 50.9 \\
        
        ORQA \cite{lee-etal-2018-ranking} & 31.3 & 45.1 \\
        
        REALM \cite{pmlr-v119-guu20a} & 40.4 & - \\
        
        DPR \cite{karpukhin-etal-2020-dense} & 41.5 & 57.9\\
        RAG \cite{10.5555/3495724.3496517} & 44.5 & 56.1\\
        DensePhrases \cite{lee-etal-2021-learning-dense} & 41.5 & 56.8\\
        PAQ \cite{lewis-etal-2021-paq} & 52.3 & -\\
        \midrule
        
        FiD* \cite{izacard-grave-2021-leveraging} (base) & 50.03 & 68.01\\
        \textbf{Ours K= 1 (with FiD base)} & \textbf{50.83} & \textbf{68.52}\\
        \textbf{Ours K= 2 (with FiD base)} & \textbf{50.94} & \textbf{68.69}\\
        \textbf{Ours K= 3 (with FiD base)} & \textbf{50.97} & \textbf{68.80}\\
        
        \midrule
        FiD* \cite{izacard-grave-2021-leveraging} (large) & 54.43 & 72.07\\

        \textbf{Ours K= 1 (with FiD large)} & \textbf{54.90} & \textbf{72.16}\\
        \textbf{Ours K= 2 (with FiD large)} & \textbf{54.99} & \textbf{72.29}\\
        \textbf{Ours K= 3 (with FiD large)} & \textbf{55.10} & \textbf{72.33}\\

    \bottomrule
    \end{tabular}
    \caption{
    Comparing EM performance of ODQA methods. \\
    * indicates the highest performance of the latest model.
    }
    \label{tab:NQ_perf_comparison_KI_1}
\end{table}

\paragraph{Comparing Overall Performance:}
We show the exact match accuracies achieved by various ODQA methods in Table \ref{tab:NQ_perf_comparison_KI_1}.
Our method achieves slightly higher performance that both FiD base and large models.
We note that this is an additional benefit of our approach, though, the primary benefit remains to be the improvement in efficiency.
The cost of FiD base model is $20.19 \times 10^{11}$ FLOPs while our system matches its accuracy at just $13.10 \times 10^{11}$ FLOPs.
We show that the performance improves further on increasing the number of knowledge iterations in our method.

\begin{figure*}[t]
\centering

    \begin{subfigure}{.28\linewidth}
        \includegraphics[width=\linewidth]{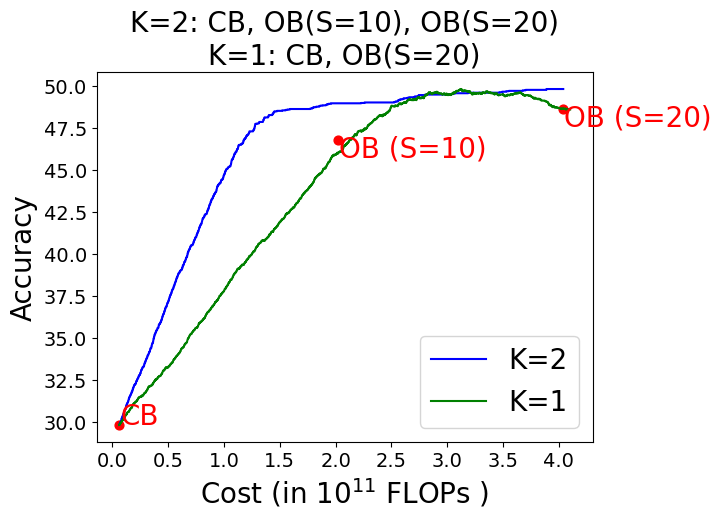}
    \end{subfigure}
    \begin{subfigure}{.28\linewidth}
          \includegraphics[width=\linewidth]{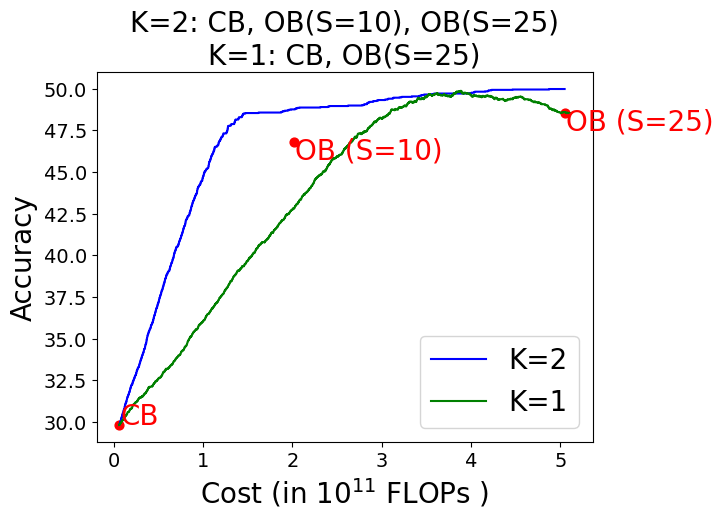}
    \end{subfigure}
    \begin{subfigure}{.28\linewidth}
         \includegraphics[width=\linewidth]{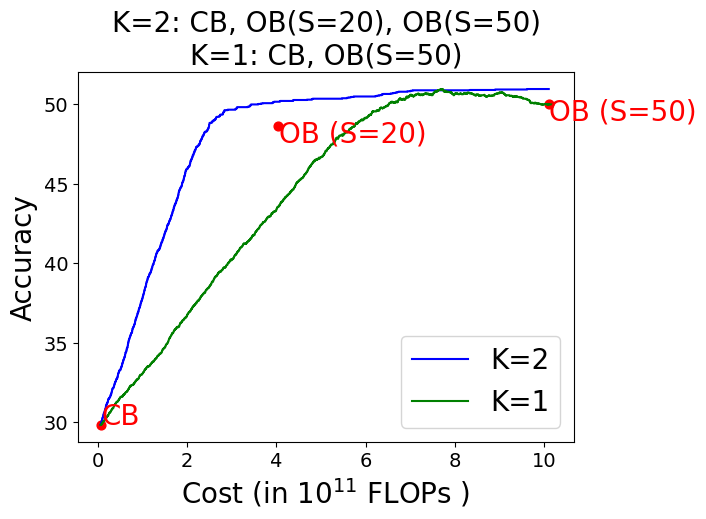}
    \end{subfigure}
   
    \caption{
    [Best viewed in color] Illustrating the \textbf{impact of multiple knowledge iterations} by plotting the accuracy-cost curves for K=1 and K=2 settings together. The system using two iterations (K=2) achieves higher AUC than its counterpart using the same amount of total knowledge (20, 25, and 50 in the three cases respectively) but with just one iteration.
    }
    \label{fig:K_comparison}    
\end{figure*}
\subsection{Two Knowledge Iterations (K = 2)}

In this setting, we use CB model and then conditionally use two knowledge iterations with the OB model. 
Specifically, if CB is not sufficiently confident then we use the OB model with $S_1$ passages ($OB_1$), and if $OB_1$ is not sufficiently confident then we use the OB model with $S_2$ passages ($OB_2$).
Figure \ref{fig:acc_cost_curves_KI_2} shows accuracy-cost curves for this setting.

\begin{table}[t]
    \centering
    \small
    \begin{tabular}{@{}lccc|ccc@{}}
        \toprule
         \textbf{CB} & \textbf{OB} & \boldmath{$S_1$} & \boldmath{$S_2$} & \textbf{$P_A$} & \textbf{$P_{PA}$}& \textbf{Baseline}\\
        \midrule

        {T5-Large} & {T5-Base} & {1} & {2} & {39.56} & \textbf{39.79} & 32.75 \\
        {T5-Large} & {T5-Base} & {2} & {5} & {42.90} & \textbf{43.14} & 36.62 \\
        {T5-Large} & {T5-Base} & {10} & {20} & {46.23} & \textbf{46.30} & 42.83 \\
        {T5-Large} & {T5-Base} & {20} & {100} & {49.04} & \textbf{49.09} & 47.07 \\
        {T5-Large} & {T5-Large} & {1} & {2} & {42.31} & \textbf{42.48} & 35.76\\
        {T5-Large} & {T5-Large} & {2} & {5} & {46.24} & \textbf{46.35} & 41.08  \\
        {T5-Large} & {T5-Large} & {10} & {20} & {49.55} & \textbf{49.71} & 46.51\\
        {T5-Large} & {T5-Large} & {20} & {100} & {52.87} & \textbf{52.97} & 51.44 \\


    \bottomrule
    \end{tabular}
    \caption{Comparing AUCs of accuracy-cost curves of the proposed and the baseline methods in K=2 setting.}
    \label{tab:auc_KI_2}
\end{table}

\paragraph{Improvement in Efficiency:}
In this setting, our method achieves larger efficiency improvements than the K=1 setting. 
For example, in case of (CB=T5-large, OB=T5-base, $S_1$=1, $S_2$=2), $OB_2$ achieves $37.48\%$ accuracy at the cost of $0.4 \times 10^{11}$ FLOPs and cascading system achieves the same accuracy at just $0.13 \times 10^{11}$ FLOPs.
We achieve similar efficiency improvements over $OB_1$ model also; in the same case, $OB_1$ achieves $32.88\%$ accuracy at $0.2 \times 10^{11}$ FLOPs and our system achieves the same accuracy at just $0.09 \times 10^{11}$ FLOPs.
Similar improvements are observed for all the configurations as we show in Appendix.

\paragraph{Improvement in Accuracy:}
As can be observed from the accuracy-cost curves, our system achieves a higher accuracy than even the $OB_2$ model. 
For example, in case of (CB=T5-large, OB=T5-base, $S_1$=10, $S_2$=20), our system achieves accuracy of $49.83\%$ that is considerably higher than the $48.61\%$ accuracy of $OB_2$ and $46.78\%$ accuracy of $OB_1$.
Furthermore, at the same cost as $OB_1$, cascading system achieves $48.98\%$ accuracy, $2.2\%$ higher than that of $OB_1$ system ($46.78\%$).
Thus, our method improves both the reader computation efficiency and the prediction accuracy.




\paragraph{Comparison with Baseline: }
In Table \ref{tab:auc_KI_2}, we show AUCs for different configurations.
Same as K=1 setting, the proposed system clearly achieves higher AUC than the baseline.

\subsection{Three Knowledge Iterations (K = 3)}
In this setting, we first use the CB model and then conditionally use three knowledge iterations with the OB model with $S_1$, $S_2$, and $S_3$ passages respectively.
Figure \ref{fig:acc_cost_curves_KI_3} shows the accuracy-cost curves for different configurations.

\paragraph{Improvement in Efficiency and Accuracy:}
We observe both efficiency and accuracy improvements in this setting also.
For example, in (CB=T5-large, OB=T5-base, $S_1$=10, $S_2$=20, $S_3$=100) configuration, $OB_3$ achieves $50.03\%$ accuracy at cost of $20.19 \times 10^{11}$ FLOPs and our system achieves the same accuracy at just $5.03 \times 10^{11}$ FLOPs.
Furthermore, our system even achieves higher accuracy than even $OB_3$.
Finally, the proposed system achieves AUC of $49.62$ which is considerably higher than that of the baseline ($47.7$).

\begin{table}[t]
    \centering
    \small
    \begin{tabular}{@{}lc@{}}
        \toprule
         \textbf{Method} & \textbf{Cost (in $10^{11}$ FLOPs)}\\
         
        \midrule
        
        FiD (base) & 20.19 \\
        Ours (at same EM as FiD base) & 3.69 \\
        \midrule
        
        FiD (large) & 70.69 \\
        Ours (at same EM as FiD large) & 20.59 \\

    \bottomrule
    \end{tabular}
    \caption{
    Comparing reader inference cost of FiD and our system at equivalent exact match accuracies on NQ.
    }
    \label{tab:efficiency_improvement}
\end{table}

\section{Impact of Knowledge Iterations}
\label{sec_impact_of_knowledge_iterations}

We demonstrate the impact of multiple knowledge iterations by plotting the accuracy-cost curves for K=1 and K=2 settings together in Figure \ref{fig:K_comparison}. 
The system using two iterations (K=2) achieves higher AUC than its counterpart using the same amount of knowledge but with just one iteration.
For the first case, we use CB and OB model (with S=20) in K=1 setting, and in the K=2 setting, we introduce an intermediate step that uses 10 passages i.e. we use CB, $OB_1$ (with $S_1$=10), and $OB_2$ (with $S_2$=20). 
The total amount of knowledge is same in both the scenarios (20 contexts) but K=2 system tries to first answer the question using just 10 passages while the K=1 system directly uses 20 passages. 
The K=2 system achieves higher AUC than K=1 (46.25 vs 43.56).
This pattern is seen in all the cases thus highlighting the positive impact of using knowledge iterations with OB model.

\section{Comparing Overall Performance}
In Table \ref{tab:NQ_perf_comparison_KI_1}, we compare the performance achieved by our system in different K settings.
With the increase in the value of K, the improvement in performance also increases.
On NQ Open, our system achieves accuracy of up to $55.10\%$ with large model and up to $50.97\%$ with base model outperforming all other reader methods such as Hard EM, ORQA, REALM, DPR, RAG, DensePhrases, PAQ, and FiD.
Similar improvements are also observed on the TriviaQA dataset.
Finally, in Table \ref{tab:efficiency_improvement}, we compare the cost of reader inference (measured in FLOPs) at equal exact match accuracies.  
We show that our method achieves considerable efficiency improvements over the FiD reader.




\section{Conclusion}

Addressing the problem of high-inference cost of reader model in open-domain QA, we explored an approach that utilizes both the `closed-book' and the `open-book' inference, and dynamically reads the external knowledge in multiple `knowledge iterations'.
Through comprehensive experiments, we demonstrated that this dynamic reading approach improves both the \textbf{inference efficiency} and the \textbf{prediction accuracy} of the reader.
Comparing with the top-performing Fusion-in-Decoder reader, this approach matches its accuracy by utilizing just $18.32\%$ of its reader inference cost (measured in FLOPs) and also outperforms it by achieving up to $55.10\%$ accuracy on NQ Open and $72.33\%$ on TQA.

\bibliography{aaai23.bib}

\begin{thebibliography}{51}
\providecommand{\natexlab}[1]{#1}

\bibitem[{Ben~Zaken, Goldberg, and Ravfogel(2022)}]{ben-zaken-etal-2022-bitfit}
Ben~Zaken, E.; Goldberg, Y.; and Ravfogel, S. 2022.
\newblock {B}it{F}it: Simple Parameter-efficient Fine-tuning for
  Transformer-based Masked Language-models.
\newblock In \emph{Proceedings of the 60th Annual Meeting of the Association
  for Computational Linguistics (Volume 2: Short Papers)}, 1--9. Dublin,
  Ireland: Association for Computational Linguistics.

\bibitem[{Chen et~al.(2017)Chen, Fisch, Weston, and
  Bordes}]{chen-etal-2017-reading}
Chen, D.; Fisch, A.; Weston, J.; and Bordes, A. 2017.
\newblock Reading {W}ikipedia to Answer Open-Domain Questions.
\newblock In \emph{Proceedings of the 55th Annual Meeting of the Association
  for Computational Linguistics (Volume 1: Long Papers)}, 1870--1879.
  Vancouver, Canada: Association for Computational Linguistics.

\bibitem[{Clark et~al.(2019)Clark, Luong, Khandelwal, Manning, and
  Le}]{clark-etal-2019-bam}
Clark, K.; Luong, M.-T.; Khandelwal, U.; Manning, C.~D.; and Le, Q.~V. 2019.
\newblock {BAM}! Born-Again Multi-Task Networks for Natural Language
  Understanding.
\newblock In \emph{Proceedings of the 57th Annual Meeting of the Association
  for Computational Linguistics}, 5931--5937. Florence, Italy: Association for
  Computational Linguistics.

\bibitem[{Das et~al.(2019)Das, Dhuliawala, Zaheer, and
  McCallum}]{das2018multistep}
Das, R.; Dhuliawala, S.; Zaheer, M.; and McCallum, A. 2019.
\newblock Multi-step Retriever-Reader Interaction for Scalable Open-domain
  Question Answering.
\newblock In \emph{International Conference on Learning Representations}.

\bibitem[{Garg and Moschitti(2021)}]{garg-moschitti-2021-will}
Garg, S.; and Moschitti, A. 2021.
\newblock Will this Question be Answered? Question Filtering via Answer Model
  Distillation for Efficient Question Answering.
\newblock In \emph{Proceedings of the 2021 Conference on Empirical Methods in
  Natural Language Processing}, 7329--7346. Online and Punta Cana, Dominican
  Republic: Association for Computational Linguistics.

\bibitem[{Goyal et~al.(2020)Goyal, Choudhury, Raje, Chakaravarthy, Sabharwal,
  and Verma}]{goyal2020power}
Goyal, S.; Choudhury, A.~R.; Raje, S.; Chakaravarthy, V.; Sabharwal, Y.; and
  Verma, A. 2020.
\newblock PoWER-BERT: Accelerating BERT inference via progressive word-vector
  elimination.
\newblock In \emph{International Conference on Machine Learning}, 3690--3699.
  PMLR.

\bibitem[{Guo, Rush, and Kim(2021)}]{guo-etal-2021-parameter}
Guo, D.; Rush, A.; and Kim, Y. 2021.
\newblock Parameter-Efficient Transfer Learning with Diff Pruning.
\newblock In \emph{Proceedings of the 59th Annual Meeting of the Association
  for Computational Linguistics and the 11th International Joint Conference on
  Natural Language Processing (Volume 1: Long Papers)}, 4884--4896. Online:
  Association for Computational Linguistics.

\bibitem[{Guu et~al.(2020)Guu, Lee, Tung, Pasupat, and
  Chang}]{pmlr-v119-guu20a}
Guu, K.; Lee, K.; Tung, Z.; Pasupat, P.; and Chang, M. 2020.
\newblock Retrieval Augmented Language Model Pre-Training.
\newblock In III, H.~D.; and Singh, A., eds., \emph{Proceedings of the 37th
  International Conference on Machine Learning}, volume 119 of
  \emph{Proceedings of Machine Learning Research}, 3929--3938. PMLR.

\bibitem[{Hou et~al.(2020)Hou, Huang, Shang, Jiang, Chen, and
  Liu}]{NEURIPS2020_6f5216f8}
Hou, L.; Huang, Z.; Shang, L.; Jiang, X.; Chen, X.; and Liu, Q. 2020.
\newblock DynaBERT: Dynamic BERT with Adaptive Width and Depth.
\newblock In Larochelle, H.; Ranzato, M.; Hadsell, R.; Balcan, M.~F.; and Lin,
  H., eds., \emph{Advances in Neural Information Processing Systems},
  volume~33, 9782--9793. Curran Associates, Inc.

\bibitem[{Houlsby et~al.(2019)Houlsby, Giurgiu, Jastrzebski, Morrone,
  De~Laroussilhe, Gesmundo, Attariyan, and Gelly}]{pmlr-v97-houlsby19a}
Houlsby, N.; Giurgiu, A.; Jastrzebski, S.; Morrone, B.; De~Laroussilhe, Q.;
  Gesmundo, A.; Attariyan, M.; and Gelly, S. 2019.
\newblock Parameter-Efficient Transfer Learning for {NLP}.
\newblock In Chaudhuri, K.; and Salakhutdinov, R., eds., \emph{Proceedings of
  the 36th International Conference on Machine Learning}, volume~97 of
  \emph{Proceedings of Machine Learning Research}, 2790--2799. PMLR.

\bibitem[{Izacard and Grave(2021)}]{izacard-grave-2021-leveraging}
Izacard, G.; and Grave, E. 2021.
\newblock Leveraging Passage Retrieval with Generative Models for Open Domain
  Question Answering.
\newblock In \emph{Proceedings of the 16th Conference of the European Chapter
  of the Association for Computational Linguistics: Main Volume}, 874--880.
  Online: Association for Computational Linguistics.

\bibitem[{Izacard et~al.(2020)Izacard, Petroni, Hosseini, De~Cao, Riedel, and
  Grave}]{izacard2020memory}
Izacard, G.; Petroni, F.; Hosseini, L.; De~Cao, N.; Riedel, S.; and Grave, E.
  2020.
\newblock A memory efficient baseline for open domain question answering.
\newblock \emph{arXiv preprint arXiv:2012.15156}.

\bibitem[{Jiao et~al.(2020)Jiao, Yin, Shang, Jiang, Chen, Li, Wang, and
  Liu}]{jiao-etal-2020-tinybert}
Jiao, X.; Yin, Y.; Shang, L.; Jiang, X.; Chen, X.; Li, L.; Wang, F.; and Liu,
  Q. 2020.
\newblock {T}iny{BERT}: Distilling {BERT} for Natural Language Understanding.
\newblock In \emph{Findings of the Association for Computational Linguistics:
  EMNLP 2020}, 4163--4174. Online: Association for Computational Linguistics.

\bibitem[{Joshi et~al.(2017)Joshi, Choi, Weld, and
  Zettlemoyer}]{joshi-etal-2017-triviaqa}
Joshi, M.; Choi, E.; Weld, D.; and Zettlemoyer, L. 2017.
\newblock {T}rivia{QA}: A Large Scale Distantly Supervised Challenge Dataset
  for Reading Comprehension.
\newblock In \emph{Proceedings of the 55th Annual Meeting of the Association
  for Computational Linguistics (Volume 1: Long Papers)}, 1601--1611.
  Vancouver, Canada: Association for Computational Linguistics.

\bibitem[{Kamath, Jia, and Liang(2020)}]{kamath-etal-2020-selective}
Kamath, A.; Jia, R.; and Liang, P. 2020.
\newblock Selective Question Answering under Domain Shift.
\newblock In \emph{Proceedings of the 58th Annual Meeting of the Association
  for Computational Linguistics}, 5684--5696. Online: Association for
  Computational Linguistics.

\bibitem[{Karpukhin et~al.(2020)Karpukhin, Oguz, Min, Lewis, Wu, Edunov, Chen,
  and Yih}]{karpukhin-etal-2020-dense}
Karpukhin, V.; Oguz, B.; Min, S.; Lewis, P.; Wu, L.; Edunov, S.; Chen, D.; and
  Yih, W.-t. 2020.
\newblock Dense Passage Retrieval for Open-Domain Question Answering.
\newblock In \emph{Proceedings of the 2020 Conference on Empirical Methods in
  Natural Language Processing (EMNLP)}, 6769--6781. Online: Association for
  Computational Linguistics.

\bibitem[{Khattab and Zaharia(2020)}]{10.1145/3397271.3401075}
Khattab, O.; and Zaharia, M. 2020.
\newblock ColBERT: Efficient and Effective Passage Search via Contextualized
  Late Interaction over BERT.
\newblock In \emph{Proceedings of the 43rd International ACM SIGIR Conference
  on Research and Development in Information Retrieval}, SIGIR '20, 39–48.
  New York, NY, USA: Association for Computing Machinery.
\newblock ISBN 9781450380164.

\bibitem[{Kim and Cho(2021)}]{kim-cho-2021-length}
Kim, G.; and Cho, K. 2021.
\newblock Length-Adaptive Transformer: Train Once with Length Drop, Use Anytime
  with Search.
\newblock In \emph{Proceedings of the 59th Annual Meeting of the Association
  for Computational Linguistics and the 11th International Joint Conference on
  Natural Language Processing (Volume 1: Long Papers)}, 6501--6511. Online:
  Association for Computational Linguistics.

\bibitem[{Kratzwald and
  Feuerriegel(2018)}]{kratzwald-feuerriegel-2018-adaptive}
Kratzwald, B.; and Feuerriegel, S. 2018.
\newblock Adaptive Document Retrieval for Deep Question Answering.
\newblock In \emph{Proceedings of the 2018 Conference on Empirical Methods in
  Natural Language Processing}, 576--581. Brussels, Belgium: Association for
  Computational Linguistics.

\bibitem[{Kwiatkowski et~al.(2019)Kwiatkowski, Palomaki, Redfield, Collins,
  Parikh, Alberti, Epstein, Polosukhin, Devlin, Lee
  et~al.}]{kwiatkowski2019natural}
Kwiatkowski, T.; Palomaki, J.; Redfield, O.; Collins, M.; Parikh, A.; Alberti,
  C.; Epstein, D.; Polosukhin, I.; Devlin, J.; Lee, K.; et~al. 2019.
\newblock Natural questions: a benchmark for question answering research.
\newblock \emph{Transactions of the Association for Computational Linguistics},
  7: 453--466.

\bibitem[{Lee et~al.(2021)Lee, Sung, Kang, and
  Chen}]{lee-etal-2021-learning-dense}
Lee, J.; Sung, M.; Kang, J.; and Chen, D. 2021.
\newblock Learning Dense Representations of Phrases at Scale.
\newblock In \emph{Proceedings of the 59th Annual Meeting of the Association
  for Computational Linguistics and the 11th International Joint Conference on
  Natural Language Processing (Volume 1: Long Papers)}, 6634--6647. Online:
  Association for Computational Linguistics.

\bibitem[{Lee et~al.(2018)Lee, Yun, Kim, Ko, and Kang}]{lee-etal-2018-ranking}
Lee, J.; Yun, S.; Kim, H.; Ko, M.; and Kang, J. 2018.
\newblock Ranking Paragraphs for Improving Answer Recall in Open-Domain
  Question Answering.
\newblock In \emph{Proceedings of the 2018 Conference on Empirical Methods in
  Natural Language Processing}, 565--569. Brussels, Belgium: Association for
  Computational Linguistics.

\bibitem[{Lewis, Denoyer, and Riedel(2019)}]{lewis-etal-2019-unsupervised}
Lewis, P.; Denoyer, L.; and Riedel, S. 2019.
\newblock Unsupervised Question Answering by Cloze Translation.
\newblock In \emph{Proceedings of the 57th Annual Meeting of the Association
  for Computational Linguistics}, 4896--4910. Florence, Italy: Association for
  Computational Linguistics.

\bibitem[{Lewis et~al.(2020)Lewis, Perez, Piktus, Petroni, Karpukhin, Goyal,
  K\"{u}ttler, Lewis, Yih, Rockt\"{a}schel, Riedel, and
  Kiela}]{10.5555/3495724.3496517}
Lewis, P.; Perez, E.; Piktus, A.; Petroni, F.; Karpukhin, V.; Goyal, N.;
  K\"{u}ttler, H.; Lewis, M.; Yih, W.-t.; Rockt\"{a}schel, T.; Riedel, S.; and
  Kiela, D. 2020.
\newblock Retrieval-Augmented Generation for Knowledge-Intensive NLP Tasks.
\newblock In \emph{Proceedings of the 34th International Conference on Neural
  Information Processing Systems}, NIPS'20. Red Hook, NY, USA: Curran
  Associates Inc.
\newblock ISBN 9781713829546.

\bibitem[{Lewis et~al.(2021)Lewis, Wu, Liu, Minervini, K{\"u}ttler, Piktus,
  Stenetorp, and Riedel}]{lewis-etal-2021-paq}
Lewis, P.; Wu, Y.; Liu, L.; Minervini, P.; K{\"u}ttler, H.; Piktus, A.;
  Stenetorp, P.; and Riedel, S. 2021.
\newblock {PAQ}: 65 Million Probably-Asked Questions and What You Can Do With
  Them.
\newblock \emph{Transactions of the Association for Computational Linguistics},
  9: 1098--1115.

\bibitem[{Li and Liang(2021)}]{li-liang-2021-prefix}
Li, X.~L.; and Liang, P. 2021.
\newblock Prefix-Tuning: Optimizing Continuous Prompts for Generation.
\newblock In \emph{Proceedings of the 59th Annual Meeting of the Association
  for Computational Linguistics and the 11th International Joint Conference on
  Natural Language Processing (Volume 1: Long Papers)}, 4582--4597. Online:
  Association for Computational Linguistics.

\bibitem[{Li et~al.(2022)Li, Wang, Tan, Nallapati, Bhatia, Arnold, Xiang, and
  Roth}]{li2022dq}
Li, Z.; Wang, Z.; Tan, M.; Nallapati, R.; Bhatia, P.; Arnold, A.; Xiang, B.;
  and Roth, D. 2022.
\newblock DQ-BART: Efficient Sequence-to-Sequence Model via Joint Distillation
  and Quantization.
\newblock \emph{arXiv preprint arXiv:2203.11239}.

\bibitem[{Luo et~al.(2022)Luo, Jain, Gupta, Einolghozati, Oguz, Chatterjee,
  Chen, Baral, and Heidari}]{luo2022study}
Luo, M.; Jain, S.; Gupta, A.; Einolghozati, A.; Oguz, B.; Chatterjee, D.; Chen,
  X.; Baral, C.; and Heidari, P. 2022.
\newblock A Study on the Efficiency and Generalization of Light Hybrid
  Retrievers.
\newblock \emph{arXiv preprint arXiv:2210.01371}.

\bibitem[{Min et~al.(2021)Min, Boyd-Graber, Alberti, Chen, Choi, Collins, Guu,
  Hajishirzi, Lee, Palomaki, Raffel, Roberts, Kwiatkowski, Lewis, Wu,
  K\"uttler, Liu, Minervini, Stenetorp, Riedel, Yang, Seo, Izacard, Petroni,
  Hosseini, Cao, Grave, Yamada, Shimaoka, Suzuki, Miyawaki, Sato, Takahashi,
  Suzuki, Fajcik, Docekal, Ondrej, Smrz, Cheng, Shen, Liu, He, Chen, Gao, Oguz,
  Chen, Karpukhin, Peshterliev, Okhonko, Schlichtkrull, Gupta, Mehdad, and
  Yih}]{pmlr-v133-min21a}
Min, S.; Boyd-Graber, J.; Alberti, C.; Chen, D.; Choi, E.; Collins, M.; Guu,
  K.; Hajishirzi, H.; Lee, K.; Palomaki, J.; Raffel, C.; Roberts, A.;
  Kwiatkowski, T.; Lewis, P.; Wu, Y.; K\"uttler, H.; Liu, L.; Minervini, P.;
  Stenetorp, P.; Riedel, S.; Yang, S.; Seo, M.; Izacard, G.; Petroni, F.;
  Hosseini, L.; Cao, N.~D.; Grave, E.; Yamada, I.; Shimaoka, S.; Suzuki, M.;
  Miyawaki, S.; Sato, S.; Takahashi, R.; Suzuki, J.; Fajcik, M.; Docekal, M.;
  Ondrej, K.; Smrz, P.; Cheng, H.; Shen, Y.; Liu, X.; He, P.; Chen, W.; Gao,
  J.; Oguz, B.; Chen, X.; Karpukhin, V.; Peshterliev, S.; Okhonko, D.;
  Schlichtkrull, M.; Gupta, S.; Mehdad, Y.; and Yih, W.-t. 2021.
\newblock NeurIPS 2020 EfficientQA Competition: Systems, Analyses and Lessons
  Learned.
\newblock In Escalante, H.~J.; and Hofmann, K., eds., \emph{Proceedings of the
  NeurIPS 2020 Competition and Demonstration Track}, volume 133 of
  \emph{Proceedings of Machine Learning Research}, 86--111. PMLR.

\bibitem[{Min et~al.(2019)Min, Chen, Hajishirzi, and
  Zettlemoyer}]{min-etal-2019-discrete}
Min, S.; Chen, D.; Hajishirzi, H.; and Zettlemoyer, L. 2019.
\newblock A Discrete Hard {EM} Approach for Weakly Supervised Question
  Answering.
\newblock In \emph{Proceedings of the 2019 Conference on Empirical Methods in
  Natural Language Processing and the 9th International Joint Conference on
  Natural Language Processing (EMNLP-IJCNLP)}, 2851--2864. Hong Kong, China:
  Association for Computational Linguistics.

\bibitem[{Mirzadeh et~al.(2020)Mirzadeh, Farajtabar, Li, Levine, Matsukawa, and
  Ghasemzadeh}]{mirzadeh2020improved}
Mirzadeh, S.~I.; Farajtabar, M.; Li, A.; Levine, N.; Matsukawa, A.; and
  Ghasemzadeh, H. 2020.
\newblock Improved knowledge distillation via teacher assistant.
\newblock In \emph{Proceedings of the AAAI Conference on Artificial
  Intelligence}, volume~34, 5191--5198.

\bibitem[{Modarressi, Mohebbi, and Pilehvar(2022)}]{modarressi2022adapler}
Modarressi, A.; Mohebbi, H.; and Pilehvar, M.~T. 2022.
\newblock AdapLeR: Speeding up Inference by Adaptive Length Reduction.
\newblock \emph{arXiv preprint arXiv:2203.08991}.

\bibitem[{Qi et~al.(2019)Qi, Lin, Mehr, Wang, and
  Manning}]{qi-etal-2019-answering}
Qi, P.; Lin, X.; Mehr, L.; Wang, Z.; and Manning, C.~D. 2019.
\newblock Answering Complex Open-domain Questions Through Iterative Query
  Generation.
\newblock In \emph{Proceedings of the 2019 Conference on Empirical Methods in
  Natural Language Processing and the 9th International Joint Conference on
  Natural Language Processing (EMNLP-IJCNLP)}, 2590--2602. Hong Kong, China:
  Association for Computational Linguistics.

\bibitem[{Raffel et~al.(2020)Raffel, Shazeer, Roberts, Lee, Narang, Matena,
  Zhou, Li, and Liu}]{JMLR:v21:20-074}
Raffel, C.; Shazeer, N.; Roberts, A.; Lee, K.; Narang, S.; Matena, M.; Zhou,
  Y.; Li, W.; and Liu, P.~J. 2020.
\newblock Exploring the Limits of Transfer Learning with a Unified Text-to-Text
  Transformer.
\newblock \emph{Journal of Machine Learning Research}, 21(140): 1--67.

\bibitem[{Roberts, Raffel, and Shazeer(2020)}]{roberts-etal-2020-much}
Roberts, A.; Raffel, C.; and Shazeer, N. 2020.
\newblock How Much Knowledge Can You Pack Into the Parameters of a Language
  Model?
\newblock In \emph{Proceedings of the 2020 Conference on Empirical Methods in
  Natural Language Processing (EMNLP)}, 5418--5426. Online: Association for
  Computational Linguistics.

\bibitem[{Rodriguez et~al.(2021)Rodriguez, Barrow, Hoyle, Lalor, Jia, and
  Boyd-Graber}]{rodriguez-etal-2021-evaluation}
Rodriguez, P.; Barrow, J.; Hoyle, A.~M.; Lalor, J.~P.; Jia, R.; and
  Boyd-Graber, J. 2021.
\newblock Evaluation Examples are not Equally Informative: How should that
  change {NLP} Leaderboards?
\newblock In \emph{Proceedings of the 59th Annual Meeting of the Association
  for Computational Linguistics and the 11th International Joint Conference on
  Natural Language Processing (Volume 1: Long Papers)}, 4486--4503. Online:
  Association for Computational Linguistics.

\bibitem[{Schick and Sch{\"u}tze(2021)}]{schick-schutze-2021-exploiting}
Schick, T.; and Sch{\"u}tze, H. 2021.
\newblock Exploiting Cloze-Questions for Few-Shot Text Classification and
  Natural Language Inference.
\newblock In \emph{Proceedings of the 16th Conference of the European Chapter
  of the Association for Computational Linguistics: Main Volume}, 255--269.
  Online: Association for Computational Linguistics.

\bibitem[{Shen et~al.(2020)Shen, Dong, Ye, Ma, Yao, Gholami, Mahoney, and
  Keutzer}]{shen2020q}
Shen, S.; Dong, Z.; Ye, J.; Ma, L.; Yao, Z.; Gholami, A.; Mahoney, M.~W.; and
  Keutzer, K. 2020.
\newblock Q-bert: Hessian based ultra low precision quantization of bert.
\newblock In \emph{Proceedings of the AAAI Conference on Artificial
  Intelligence}, volume~34, 8815--8821.

\bibitem[{Tao et~al.(2022)Tao, Hou, Zhang, Shang, Jiang, Liu, Luo, and
  Wong}]{tao2022compression}
Tao, C.; Hou, L.; Zhang, W.; Shang, L.; Jiang, X.; Liu, Q.; Luo, P.; and Wong,
  N. 2022.
\newblock Compression of Generative Pre-trained Language Models via
  Quantization.
\newblock \emph{arXiv preprint arXiv:2203.10705}.

\bibitem[{Varshney et~al.(2022)Varshney, Banerjee, Gokhale, and
  Baral}]{varshney-etal-2022-unsupervised}
Varshney, N.; Banerjee, P.; Gokhale, T.; and Baral, C. 2022.
\newblock Unsupervised Natural Language Inference Using {PHL} Triplet
  Generation.
\newblock In \emph{Findings of the Association for Computational Linguistics:
  ACL 2022}, 2003--2016. Dublin, Ireland: Association for Computational
  Linguistics.

\bibitem[{Varshney and Baral(2022)}]{varshney2022model}
Varshney, N.; and Baral, C. 2022.
\newblock Model Cascading: Towards Jointly Improving Efficiency and Accuracy of
  NLP Systems.
\newblock \emph{arXiv preprint arXiv:2210.05528}.

\bibitem[{Varshney, Mishra, and
  Baral(2022{\natexlab{a}})}]{varshney-etal-2022-ildae}
Varshney, N.; Mishra, S.; and Baral, C. 2022{\natexlab{a}}.
\newblock {ILDAE}: Instance-Level Difficulty Analysis of Evaluation Data.
\newblock In \emph{Proceedings of the 60th Annual Meeting of the Association
  for Computational Linguistics (Volume 1: Long Papers)}, 3412--3425. Dublin,
  Ireland: Association for Computational Linguistics.

\bibitem[{Varshney, Mishra, and
  Baral(2022{\natexlab{b}})}]{varshney-etal-2022-investigating}
Varshney, N.; Mishra, S.; and Baral, C. 2022{\natexlab{b}}.
\newblock Investigating Selective Prediction Approaches Across Several Tasks in
  {IID}, {OOD}, and Adversarial Settings.
\newblock In \emph{Findings of the Association for Computational Linguistics:
  ACL 2022}, 1995--2002. Dublin, Ireland: Association for Computational
  Linguistics.

\bibitem[{Varshney, Mishra, and
  Baral(2022{\natexlab{c}})}]{varshney-etal-2022-towards}
Varshney, N.; Mishra, S.; and Baral, C. 2022{\natexlab{c}}.
\newblock Towards Improving Selective Prediction Ability of {NLP} Systems.
\newblock In \emph{Proceedings of the 7th Workshop on Representation Learning
  for NLP}, 221--226. Dublin, Ireland: Association for Computational
  Linguistics.

\bibitem[{Wang, Wohlwend, and Lei(2020)}]{wang-etal-2020-structured}
Wang, Z.; Wohlwend, J.; and Lei, T. 2020.
\newblock Structured Pruning of Large Language Models.
\newblock In \emph{Proceedings of the 2020 Conference on Empirical Methods in
  Natural Language Processing (EMNLP)}, 6151--6162. Online: Association for
  Computational Linguistics.

\bibitem[{Wang et~al.(2021)Wang, Yu, Firat, and Cao}]{Wang2021TowardsZL}
Wang, Z.; Yu, A.~W.; Firat, O.; and Cao, Y. 2021.
\newblock Towards Zero-Label Language Learning.
\newblock \emph{ArXiv}, abs/2109.09193.

\bibitem[{Xin et~al.(2020)Xin, Tang, Lee, Yu, and Lin}]{xin-etal-2020-deebert}
Xin, J.; Tang, R.; Lee, J.; Yu, Y.; and Lin, J. 2020.
\newblock {D}ee{BERT}: Dynamic Early Exiting for Accelerating {BERT} Inference.
\newblock In \emph{Proceedings of the 58th Annual Meeting of the Association
  for Computational Linguistics}, 2246--2251. Online: Association for
  Computational Linguistics.

\bibitem[{Xin et~al.(2021)Xin, Tang, Yu, and Lin}]{xin-etal-2021-art}
Xin, J.; Tang, R.; Yu, Y.; and Lin, J. 2021.
\newblock The Art of Abstention: Selective Prediction and Error Regularization
  for Natural Language Processing.
\newblock In \emph{Proceedings of the 59th Annual Meeting of the Association
  for Computational Linguistics and the 11th International Joint Conference on
  Natural Language Processing (Volume 1: Long Papers)}, 1040--1051. Online:
  Association for Computational Linguistics.

\bibitem[{Yamada, Asai, and Hajishirzi(2021)}]{yamada-etal-2021-efficient}
Yamada, I.; Asai, A.; and Hajishirzi, H. 2021.
\newblock Efficient Passage Retrieval with Hashing for Open-domain Question
  Answering.
\newblock In \emph{Proceedings of the 59th Annual Meeting of the Association
  for Computational Linguistics and the 11th International Joint Conference on
  Natural Language Processing (Volume 2: Short Papers)}, 979--986. Online:
  Association for Computational Linguistics.

\bibitem[{Zhang et~al.(2020)Zhang, Hou, Yin, Shang, Chen, Jiang, and
  Liu}]{zhang-etal-2020-ternarybert}
Zhang, W.; Hou, L.; Yin, Y.; Shang, L.; Chen, X.; Jiang, X.; and Liu, Q. 2020.
\newblock {T}ernary{BERT}: Distillation-aware Ultra-low Bit {BERT}.
\newblock In \emph{Proceedings of the 2020 Conference on Empirical Methods in
  Natural Language Processing (EMNLP)}, 509--521. Online: Association for
  Computational Linguistics.

\bibitem[{Zhao, Lu, and Lee(2021)}]{zhao-etal-2021-sparta}
Zhao, T.; Lu, X.; and Lee, K. 2021.
\newblock {SPARTA}: Efficient Open-Domain Question Answering via Sparse
  Transformer Matching Retrieval.
\newblock In \emph{Proceedings of the 2021 Conference of the North American
  Chapter of the Association for Computational Linguistics: Human Language
  Technologies}, 565--575. Online: Association for Computational Linguistics.

\end{thebibliography}

\clearpage

\section*{Appendix}
\appendix

\section{One Knowledge Iteration (K=1)}
\label{sec_appendix_K_1}
\begin{figure*}[t]
\centering

    \begin{subfigure}{.28\linewidth}
        \includegraphics[width=\linewidth]{Pictures/results_multiplication_of_all_tokens/KI_1/nqopen_t5_large_ssm_nq_predictions___nq_reader_base_1_contexts.png}
    \end{subfigure}
    \begin{subfigure}{.28\linewidth}
         \includegraphics[width=\linewidth]{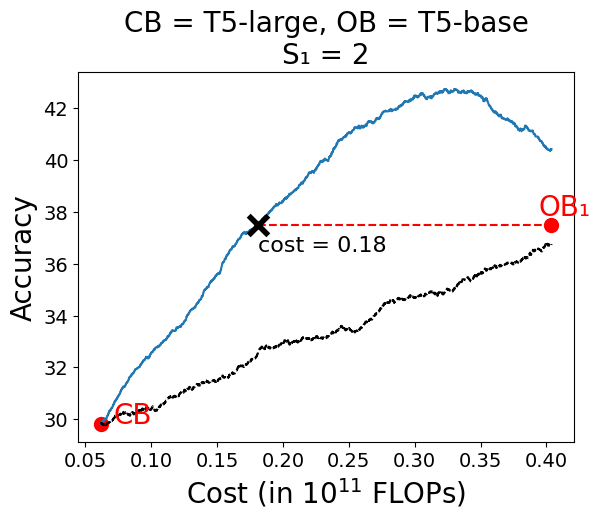}
    \end{subfigure}
    \begin{subfigure}{.28\linewidth}
         \includegraphics[width=\linewidth]{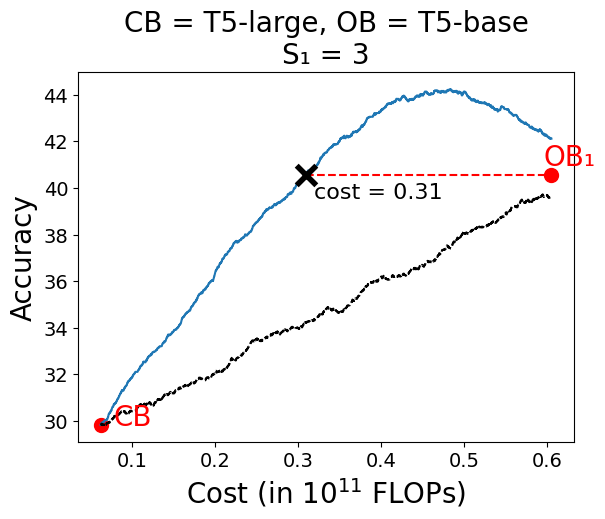}
    \end{subfigure}
    \begin{subfigure}{.28\linewidth}
         \includegraphics[width=\linewidth]{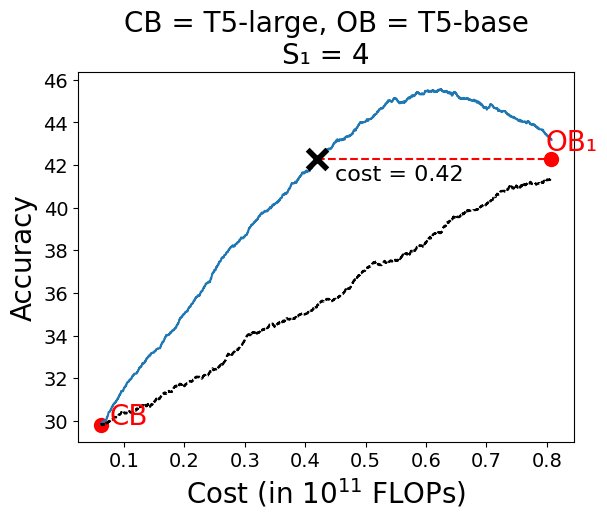}
    \end{subfigure}
    \begin{subfigure}{.28\linewidth}
         \includegraphics[width=\linewidth]{Pictures/results_multiplication_of_all_tokens/KI_1/nqopen_t5_large_ssm_nq_predictions___nq_reader_base_5_contexts.png}
    \end{subfigure}
    \begin{subfigure}{.28\linewidth}
         \includegraphics[width=\linewidth]{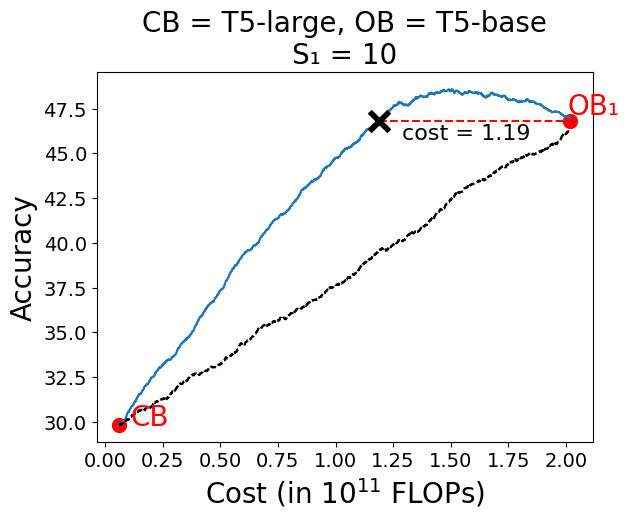}
    \end{subfigure}
    \begin{subfigure}{.28\linewidth}
         \includegraphics[width=\linewidth]{Pictures/results_multiplication_of_all_tokens/KI_1/nqopen_t5_large_ssm_nq_predictions___nq_reader_base_20_contexts.png}
    \end{subfigure}
    \begin{subfigure}{.28\linewidth}
         \includegraphics[width=\linewidth]{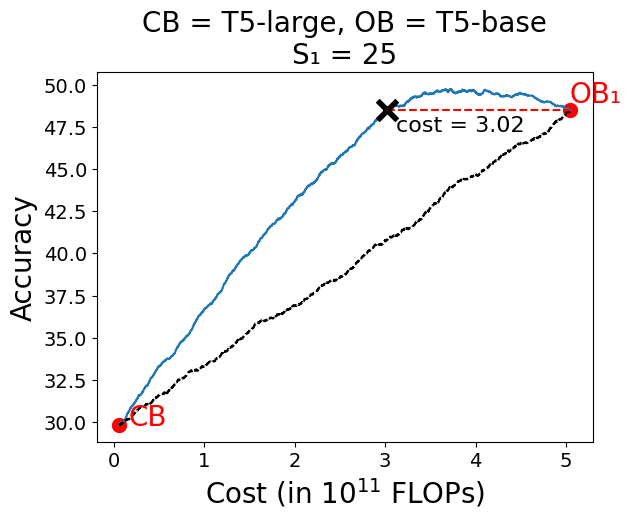}
    \end{subfigure}
    \begin{subfigure}{.28\linewidth}
         \includegraphics[width=\linewidth]{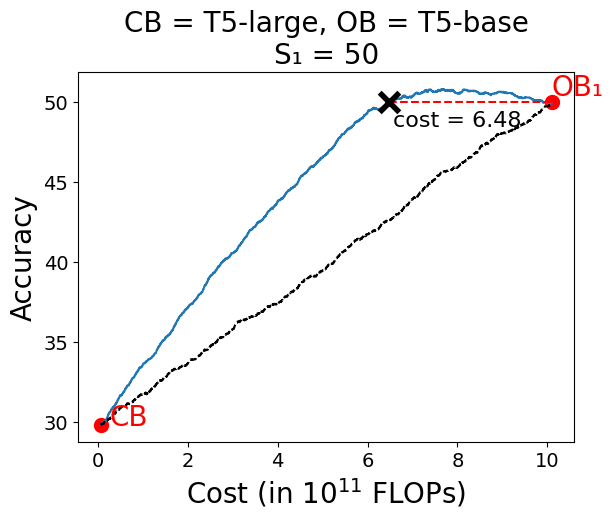}
    \end{subfigure}
    \caption{Accuracy-cost curves of the proposed cascading system (in blue) and baseline system (in black) for K=1 setting on NQ. 
    Red points correspond to the accuracy and cost values of the individual models $CB$ and $OB_1$ (leveraging $S_1$ knowledge statements). 
    Point of intersection of red dashed line drawn from $OB_1$ on the blue curve correspond to cost at which the cascading system achieves the same accuracy as $OB_1$. 
    }
    \label{fig:supp_acc_cost_curves_KI_1_base}    
\end{figure*}

\begin{figure*}[t]
\centering

    \begin{subfigure}{.28\linewidth}
        \includegraphics[width=\linewidth]{Pictures/results_multiplication_of_all_tokens/KI_1/nqopen_t5_large_ssm_nq_predictions___nq_reader_large_1_contexts.png}
    \end{subfigure}
    \begin{subfigure}{.28\linewidth}
         \includegraphics[width=\linewidth]{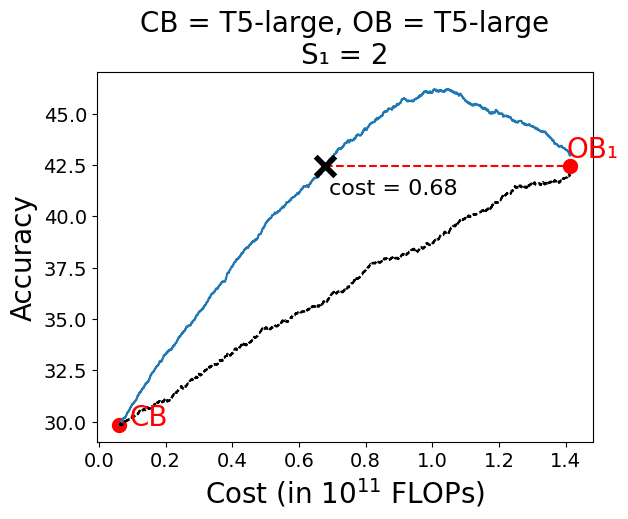}
    \end{subfigure}
    \begin{subfigure}{.28\linewidth}
         \includegraphics[width=\linewidth]{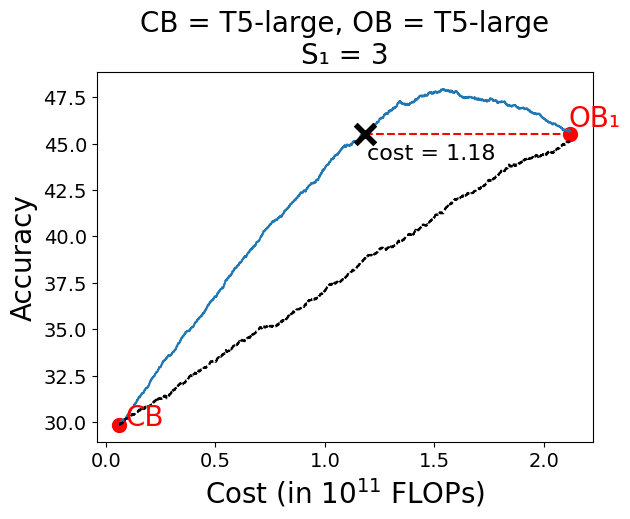}
    \end{subfigure}
    \begin{subfigure}{.28\linewidth}
         \includegraphics[width=\linewidth]{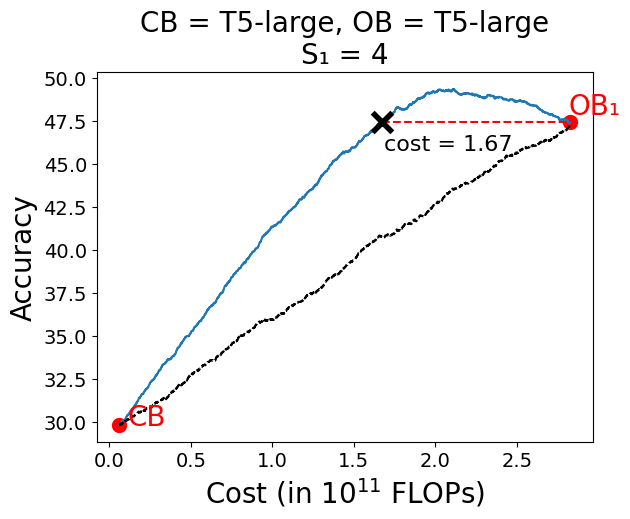}
    \end{subfigure}
    \begin{subfigure}{.28\linewidth}
         \includegraphics[width=\linewidth]{Pictures/results_multiplication_of_all_tokens/KI_1/nqopen_t5_large_ssm_nq_predictions___nq_reader_large_5_contexts.png}
    \end{subfigure}
    \begin{subfigure}{.28\linewidth}
         \includegraphics[width=\linewidth]{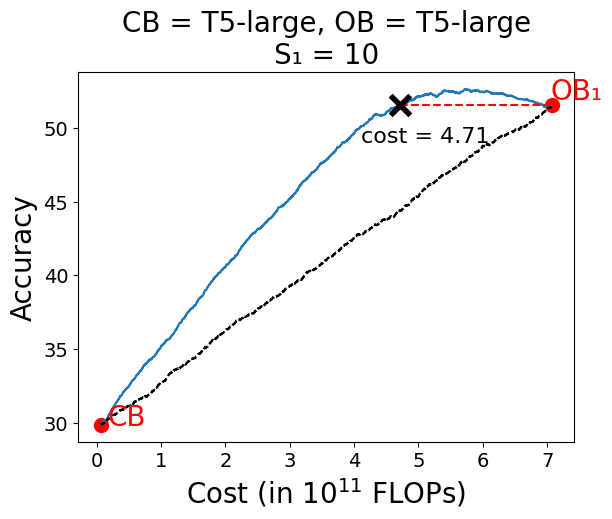}
    \end{subfigure}
    \begin{subfigure}{.28\linewidth}
         \includegraphics[width=\linewidth]{Pictures/results_multiplication_of_all_tokens/KI_1/nqopen_t5_large_ssm_nq_predictions___nq_reader_large_20_contexts.png}
    \end{subfigure}
    \begin{subfigure}{.28\linewidth}
         \includegraphics[width=\linewidth]{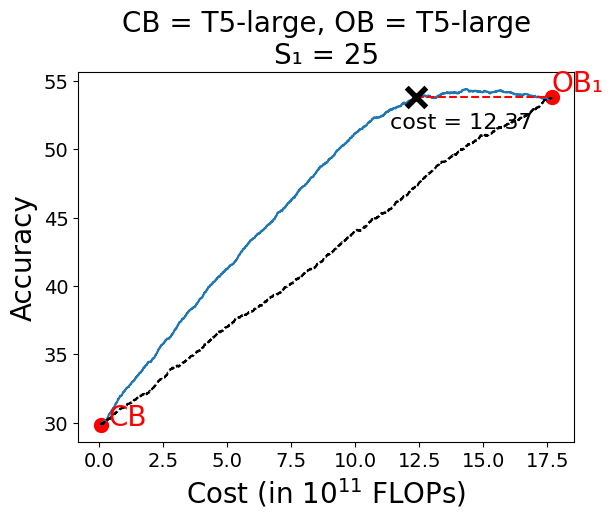}
    \end{subfigure}
    \begin{subfigure}{.28\linewidth}
         \includegraphics[width=\linewidth]{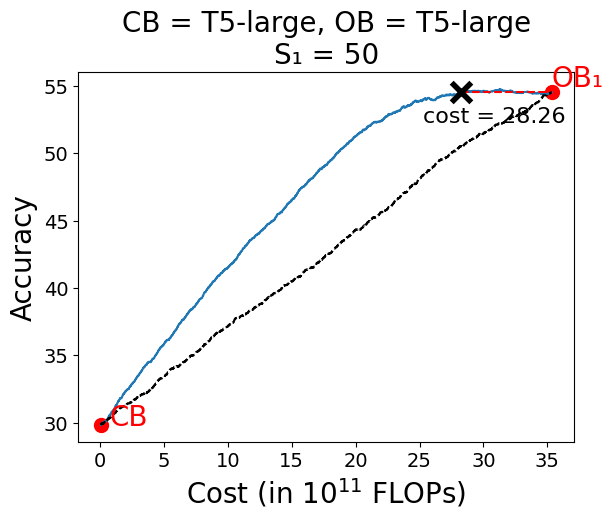}
    \end{subfigure}
    \caption{Accuracy-cost curves of the proposed cascading system (in blue) and baseline system (in black) for K=1 setting on NQ. 
    Red points correspond to the accuracy and cost values of the individual models $CB$ and $OB_1$ (leveraging $S_1$ knowledge statements). 
    Point of intersection of red dashed line drawn from $OB_1$ on the blue curve correspond to cost at which the cascading system achieves the same accuracy as $OB_1$. 
    }
    \label{fig:supp_acc_cost_curves_KI_1_large}    
\end{figure*}

In this setting, we first use the CB model and if it is not sufficiently confident then we use the OB model with $S_1$ knowledge statements.
Figure \ref{fig:supp_acc_cost_curves_KI_1_base} and \ref{fig:supp_acc_cost_curves_KI_1_large} show accuracy-cost curves of different configurations for this setting.

\begin{table*}[]
\small
\resizebox{\textwidth}{!}{%
\begin{tabular}{l|llllllllll}
\toprule

\textbf{Method $\backslash$ \boldmath{$S_1$}} &
\textbf{1} &  \textbf{2} &  \textbf{3} &  \textbf{4} &  \textbf{5} &  \textbf{10} &  \textbf{20} &  \textbf{25} &  \textbf{50} &  \textbf{100} \\

\midrule
    
    \textbf{Random} & 30.96 & 33.28 & 34.73 & 35.74 & 36.05 & 38.12 & 39.27 & 39.14 & 39.96 & 39.94\\
    \textbf{Heuristic} & 31.90 & 33.97 & 35.62 & 36.32 & 36.90 & 38.77 & 39.69 & 39.62 & 40.22 & 40.28\\
    \midrule
     \boldmath{$P_{PA}$} & \textbf{36.61} & \textbf{38.53} & \textbf{39.64} & \textbf{40.59} & \textbf{41.14} & \textbf{42.82} & \textbf{43.69} & \textbf{43.69} & \textbf{44.43} & \textbf{44.29} \\
    
    \boldmath{$P_F$} & 36.15 & 38.09 & 39.16 & 40.15 & 40.77 & 42.47 & 43.34 & 43.36 & 44.13 & 43.95 \\
    \boldmath{$P_{FL}$} & 36.13 & 38.08 & 39.15 & 40.16 & 40.76 & 42.46 & 43.35 & 43.36 & 44.16 & 43.97 \\
    \boldmath{$P_A$} & {36.39} & {38.35} & {39.49} & {40.45} & {41.00} & {42.74} & {43.56} & {43.54} & {44.30} & {44.15} \\

\bottomrule
\end{tabular}
}
\caption{Comparing AUCs of accuracy-cost curves of different cascading techniques for (CB=T5-large and OB=T5-base) configuration.}
\label{tab:auc_full_K_1_base}

\end{table*}

\begin{table*}[]
\small
\resizebox{\textwidth}{!}{%
\begin{tabular}{l|llllllllll}
\toprule

\textbf{Method $\backslash$ \boldmath{$S_1$}} &
\textbf{1} &  \textbf{2} &  \textbf{3} &  \textbf{4} &  \textbf{5} &  \textbf{10} &  \textbf{20} &  \textbf{25} &  \textbf{50} &  \textbf{100} \\

\midrule
    
    \textbf{Random} & 
    33.14 & 36.4 & 37.81 & 38.87 & 39.47 & 40.92 & 41.91 & 42.21 & 42.54 & 42.62\\
    \midrule
     \boldmath{$P_{PA}$} & \textbf{38.45} & \textbf{41.07} & \textbf{42.29} & \textbf{43.18} & \textbf{43.76} & \textbf{45.40} & \textbf{46.34} & \textbf{46.52} & \textbf{46.71} & \textbf{46.77} \\
    \boldmath{$P_F$} & 38.00 & 40.68 & 41.92 & 42.87 & 43.47 & 45.11 & 46.09 & 46.31 & 46.45 & 46.46 \\
    \boldmath{$P_{FL}$} & 37.96 & 40.65 & 41.89 & 42.85 & 43.44 & 45.09 & 46.06 & 46.28 & 46.44 & 46.46 \\
    \boldmath{$P_A$} & {38.26} & {40.93} & {42.17} & {43.08} & {43.66} & {45.28} & {46.25} & {46.43} & {46.61} & {46.69} \\

\bottomrule
\end{tabular}
}
\caption{Comparing AUCs of accuracy-cost curves of different cascading techniques for (CB=T5-large and OB=T5-large) configuration.}
\label{tab:auc_full_K_1_large}

\end{table*}

\paragraph{Improvement in Efficiency:}
These curves show that the cascading system matches the accuracy of the OB model at a much lesser computation cost. 
This cost value corresponds to the point of intersection on the curve with a straight horizontal line drawn from $OB_1$ (red dashed line).
For example, in the case of (CB= T5-large, OB=T5-base, $S_1=5$), $OB_1$ achieves $43.30\%$ accuracy at the computation cost of $1.01 \times 10^{11}$ FLOPs while the cascading system achieves the same accuracy at the cost of just $0.54 \times 10^{11}$ FLOPs and 
in the case of (CB= T5-large, OB=T5-base, $S_1=10$), $OB_1$ achieves $46.79\%$ accuracy at the computation cost of $2.02 \times 10^{11}$ FLOPs while the cascading system achieves the same accuracy at the cost of just $1.18 \times 10^{11}$ FLOPs.
Such improvements are observed for all the cases.
This efficiency benefit comes from using the low-cost closed-book model for some instances where it is likely to be correct and using the more expensive open-book model only for the other instances.

\paragraph{Comparison of Methods:}
From the accuracy-cost curves, it is clear that our cascading method that uses $P_A$ as the prediction confidence (blue curve) outperforms the random baseline (black curve) as it has a higher area under its curve.
The demonstrates the effectiveness of the way we compute the confidence scores and utilize them to build a cascading reader.
In Table \ref{tab:auc_full_K_1_base} and \ref{tab:auc_full_K_1_large}, we compare the AUC values achieved by baseline and different cascading methods using $P_A$, $P_F$, and $P_{FL}$ as prediction confidences. 
All the approaches considerably higher AUCs than the baseline.
$P_A$ achieves the highest AUC in all the cases.

\section{Two Knowledge Iterations (K=2)}
\label{sec_appendix_K_2}
\begin{figure*}[t]
\centering

    \begin{subfigure}{.28\linewidth}
        \includegraphics[width=\linewidth]{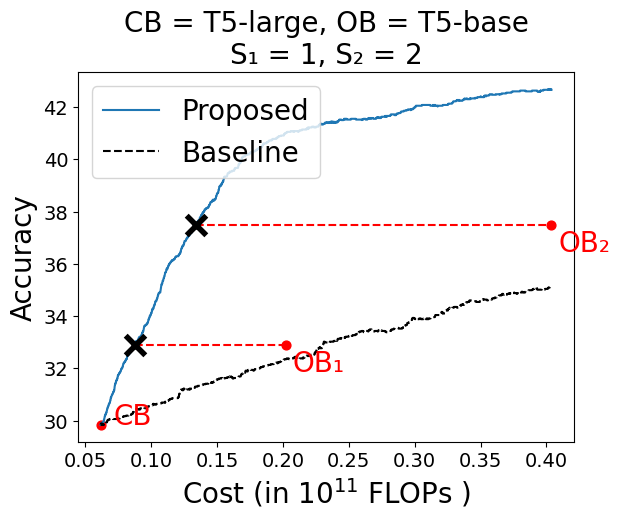}
    \end{subfigure}
    \begin{subfigure}{.28\linewidth}
         \includegraphics[width=\linewidth]{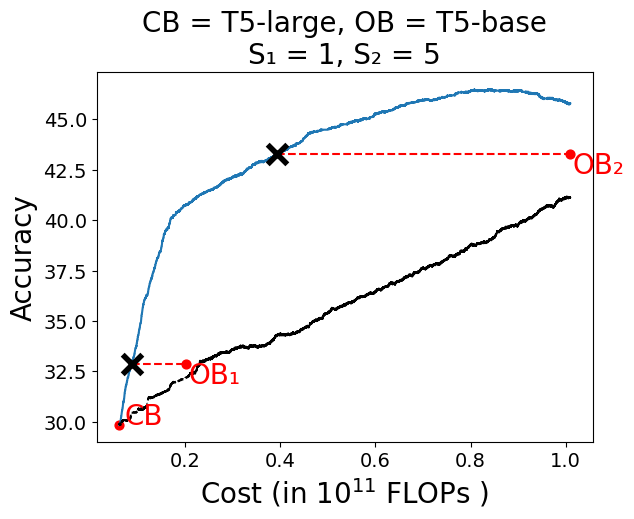}
    \end{subfigure}
    \begin{subfigure}{.28\linewidth}
         \includegraphics[width=\linewidth]{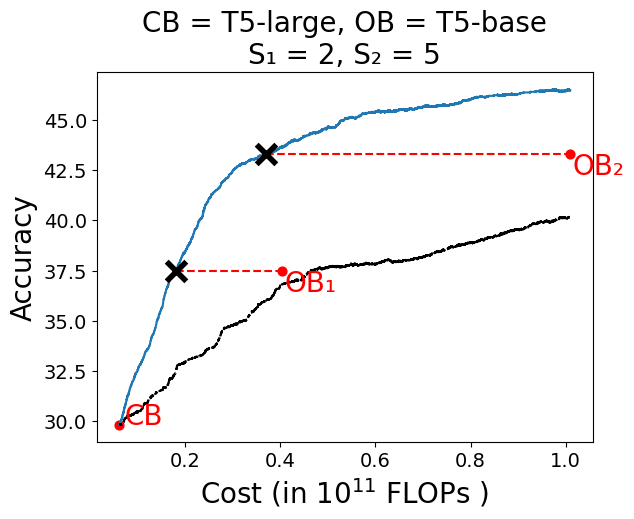}
    \end{subfigure}
    \begin{subfigure}{.28\linewidth}
         \includegraphics[width=\linewidth]{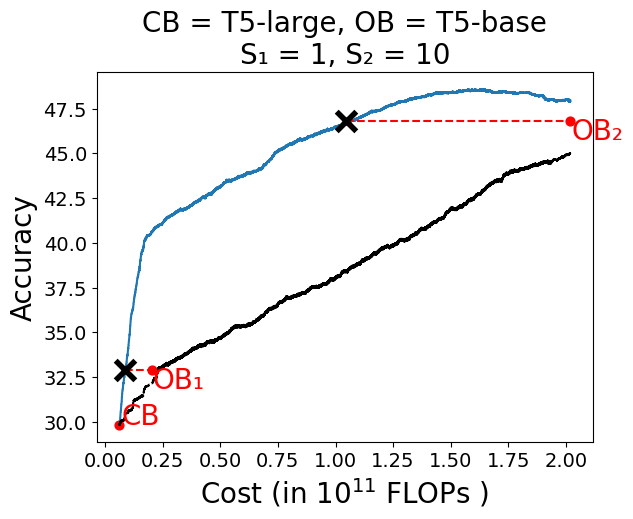}
    \end{subfigure}
    \begin{subfigure}{.28\linewidth}
         \includegraphics[width=\linewidth]{Pictures/results_multiplication_of_all_tokens/KI_2/nqopen_t5_large_ssm_nq_predictions___nq_reader_base_2_contexts___nq_reader_base_5_contexts.png}
    \end{subfigure}
    \begin{subfigure}{.28\linewidth}
         \includegraphics[width=\linewidth]{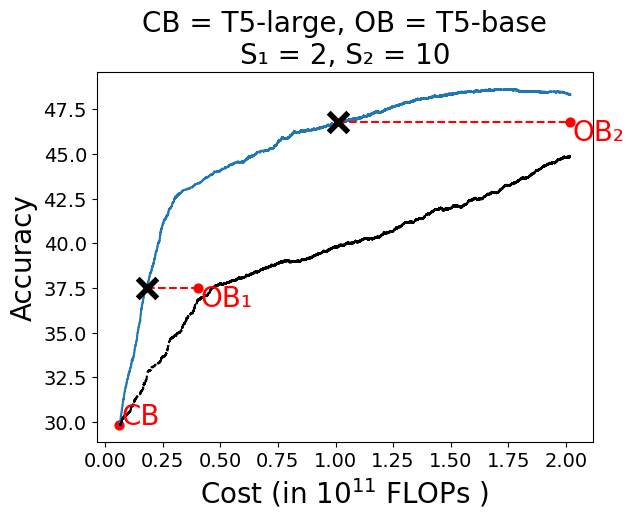}
    \end{subfigure}
    \begin{subfigure}{.28\linewidth}
         \includegraphics[width=\linewidth]{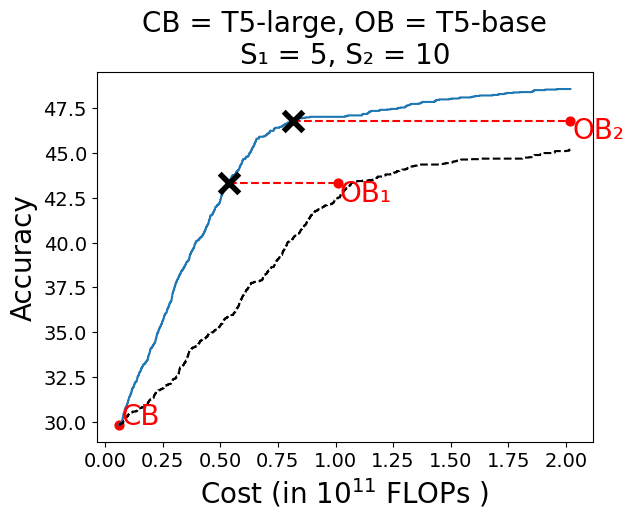}
    \end{subfigure}
    \begin{subfigure}{.28\linewidth}
         \includegraphics[width=\linewidth]{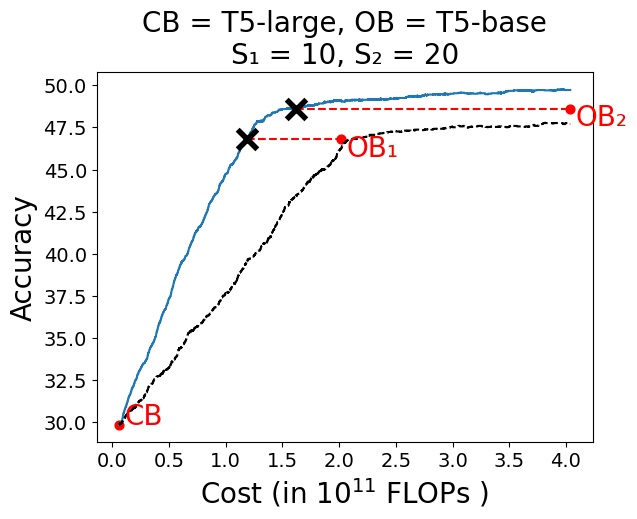}
    \end{subfigure}
    \begin{subfigure}{.28\linewidth}
         \includegraphics[width=\linewidth]{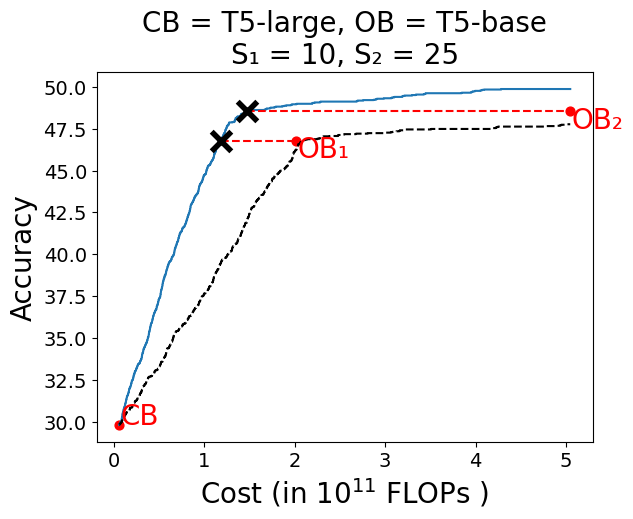}
    \end{subfigure}
    \begin{subfigure}{.28\linewidth}
         \includegraphics[width=\linewidth]{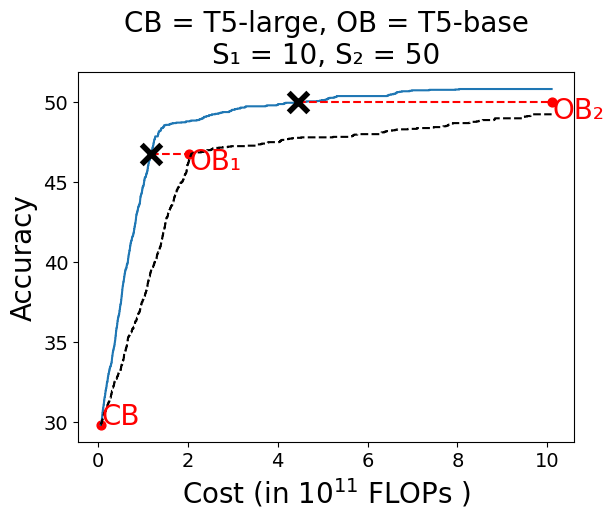}
    \end{subfigure}
    \begin{subfigure}{.28\linewidth}
         \includegraphics[width=\linewidth]{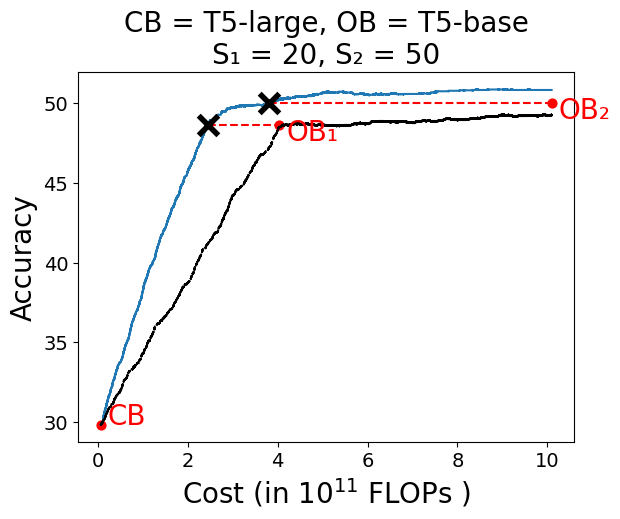}
    \end{subfigure}
    \begin{subfigure}{.28\linewidth}
         \includegraphics[width=\linewidth]{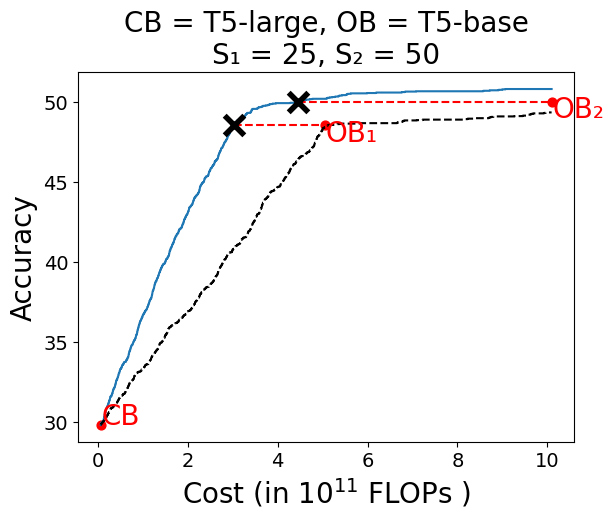}
    \end{subfigure}
    \begin{subfigure}{.28\linewidth}
         \includegraphics[width=\linewidth]{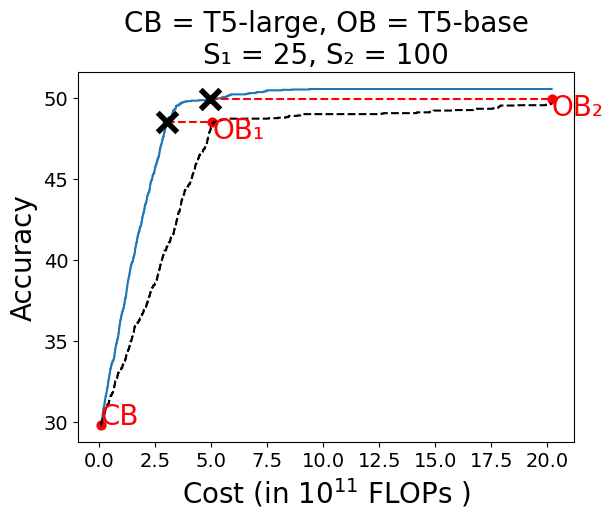}
    \end{subfigure}
    \begin{subfigure}{.28\linewidth}
         \includegraphics[width=\linewidth]{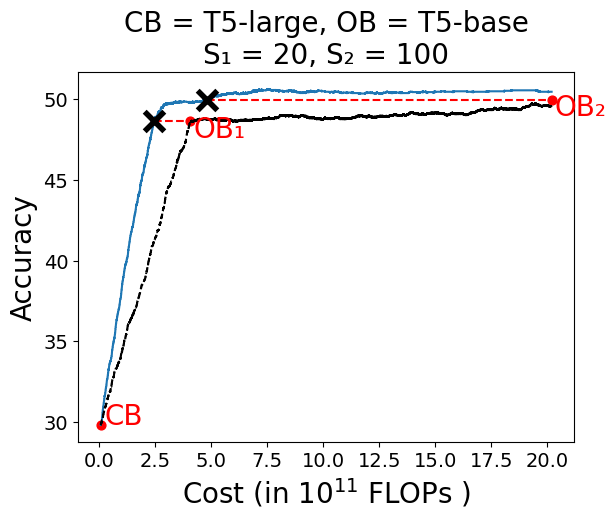}
    \end{subfigure}
    \begin{subfigure}{.28\linewidth}
         \includegraphics[width=\linewidth]{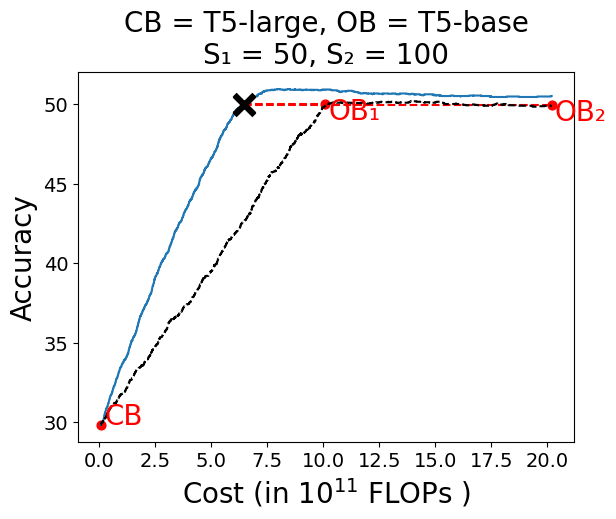}
    \end{subfigure}

    \caption{
    Accuracy-cost curves of the proposed cascading system (in blue) and baseline system (in black) for K=2 setting on NQ. 
    Red points correspond to the accuracy and cost values of the individual models $CB$, $OB_1$ (leveraging $S_1$ knowledge statements), and $OB_2$ (leveraging $S_2$ knowledge statements). 
    Points of intersection of red dashed lines drawn from $OB_1$ and $OB_2$ on the blue curve correspond to costs at which the cascading system achieves the same accuracy as $OB_1$ and $OB_2$ respectively. $OB_1$ and $OB_2$ are the same models but differ only in the $S$ value.
    }
    \label{fig:supp_acc_cost_curves_KI_2_base}    
\end{figure*}

\begin{figure*}[t]
\centering

    \begin{subfigure}{.28\linewidth}
        \includegraphics[width=\linewidth]{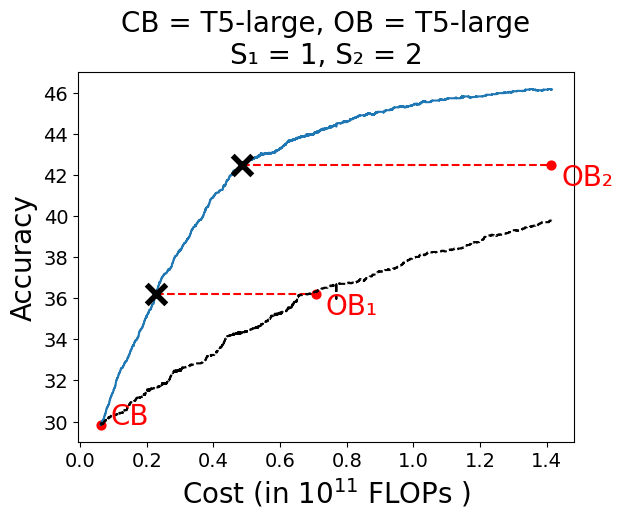}
    \end{subfigure}
    \begin{subfigure}{.28\linewidth}
         \includegraphics[width=\linewidth]{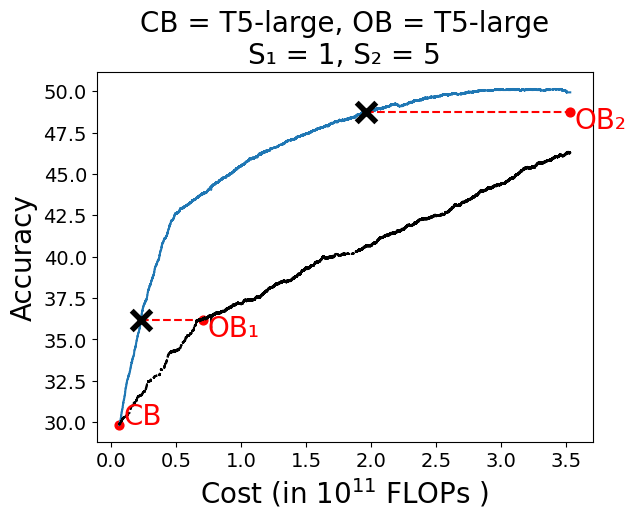}
    \end{subfigure}
    \begin{subfigure}{.28\linewidth}
         \includegraphics[width=\linewidth]{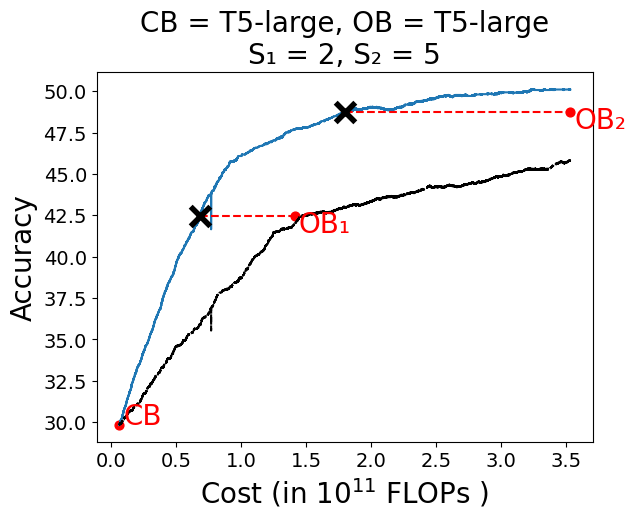}
    \end{subfigure}
    \begin{subfigure}{.28\linewidth}
         \includegraphics[width=\linewidth]{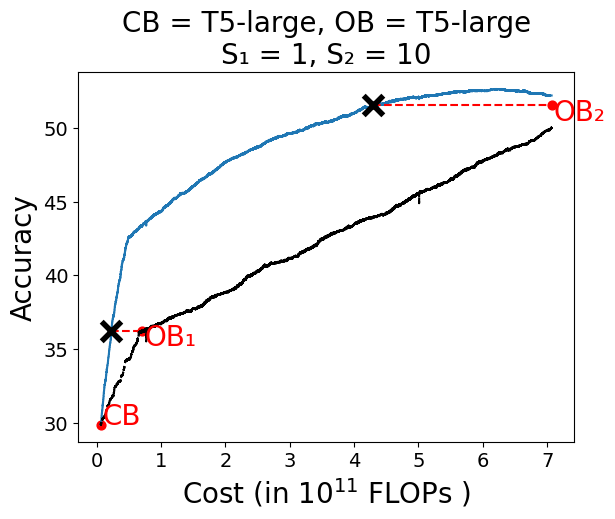}
    \end{subfigure}
    \begin{subfigure}{.28\linewidth}
         \includegraphics[width=\linewidth]{Pictures/results_multiplication_of_all_tokens/KI_2/nqopen_t5_large_ssm_nq_predictions___nq_reader_large_2_contexts___nq_reader_large_5_contexts.png}
    \end{subfigure}
    \begin{subfigure}{.28\linewidth}
         \includegraphics[width=\linewidth]{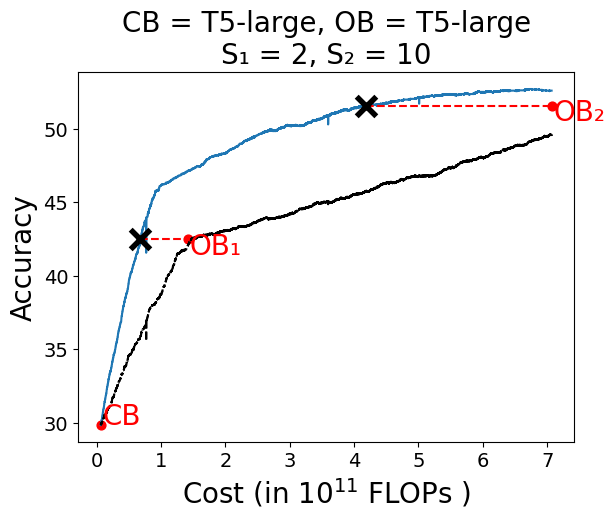}
    \end{subfigure}
    \begin{subfigure}{.28\linewidth}
         \includegraphics[width=\linewidth]{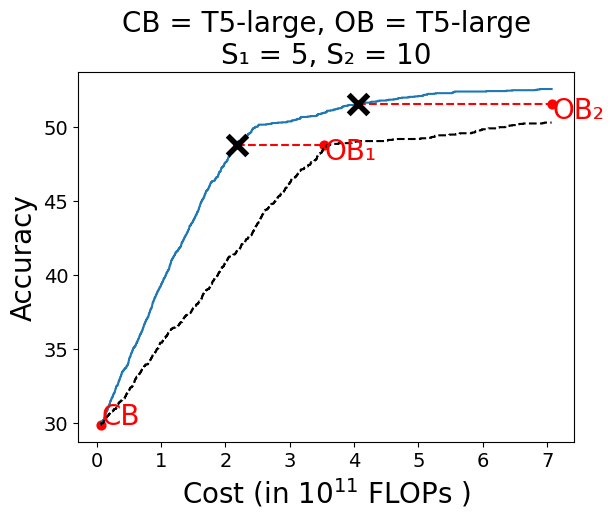}
    \end{subfigure}
    \begin{subfigure}{.28\linewidth}
         \includegraphics[width=\linewidth]{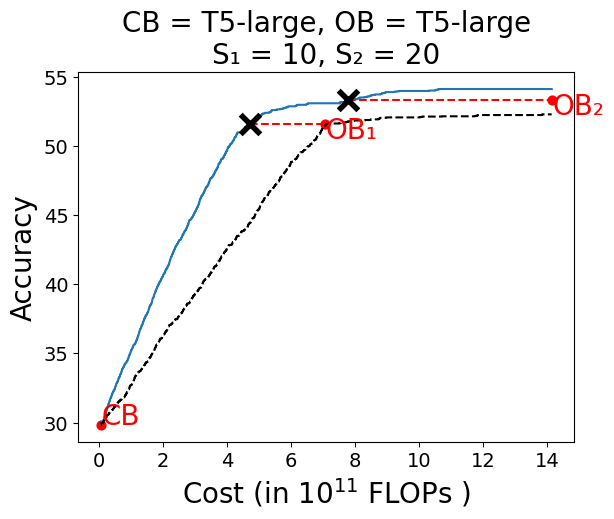}
    \end{subfigure}
    \begin{subfigure}{.28\linewidth}
         \includegraphics[width=\linewidth]{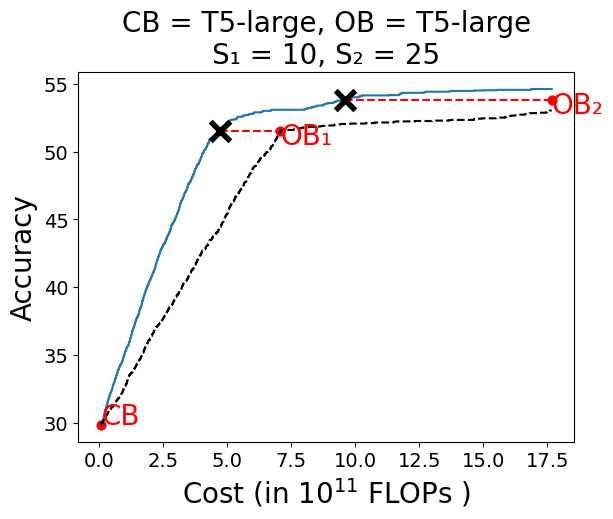}
    \end{subfigure}
    \begin{subfigure}{.28\linewidth}
         \includegraphics[width=\linewidth]{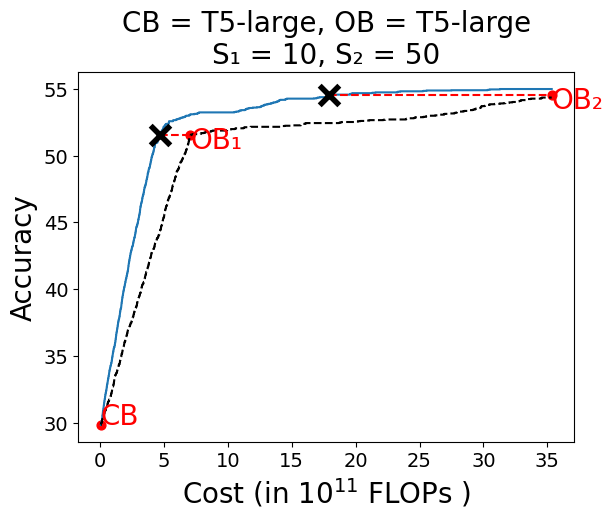}
    \end{subfigure}
    \begin{subfigure}{.28\linewidth}
         \includegraphics[width=\linewidth]{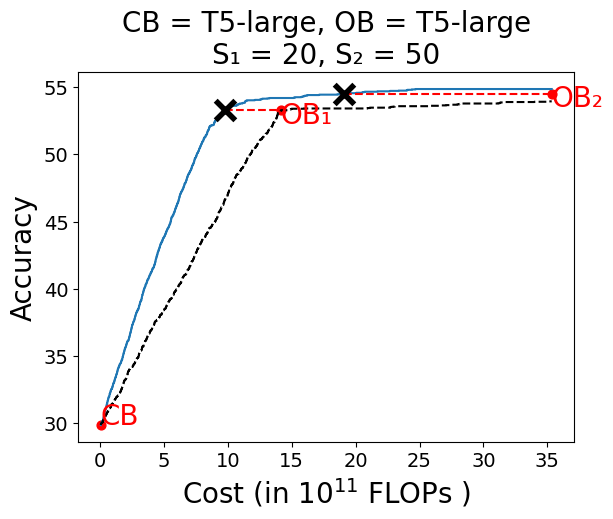}
    \end{subfigure}
    \begin{subfigure}{.28\linewidth}
         \includegraphics[width=\linewidth]{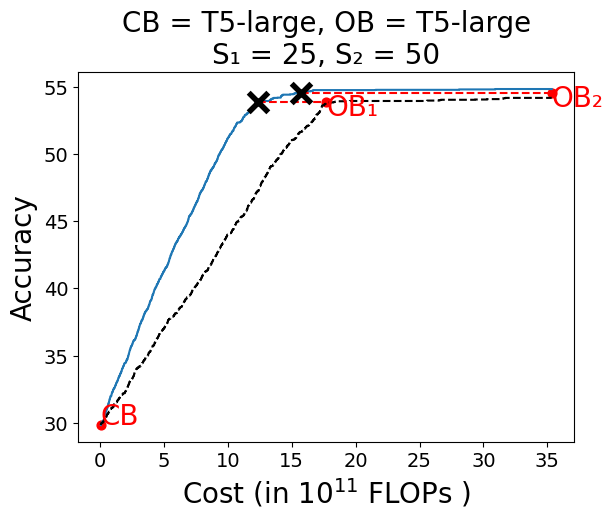}
    \end{subfigure}
    \begin{subfigure}{.28\linewidth}
         \includegraphics[width=\linewidth]{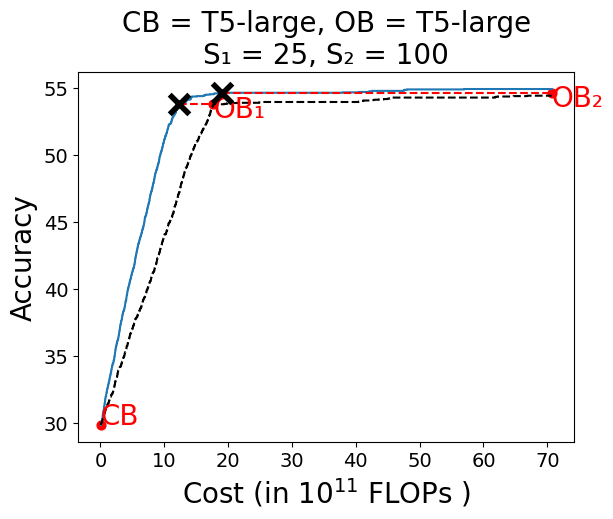}
    \end{subfigure}
    \begin{subfigure}{.28\linewidth}
         \includegraphics[width=\linewidth]{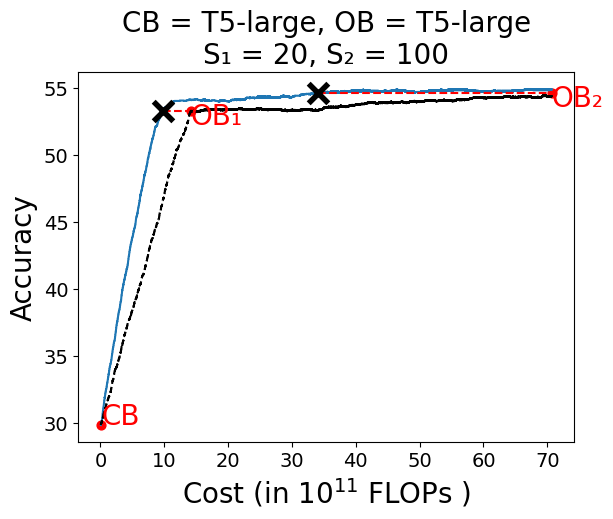}
    \end{subfigure}
    \begin{subfigure}{.28\linewidth}
         \includegraphics[width=\linewidth]{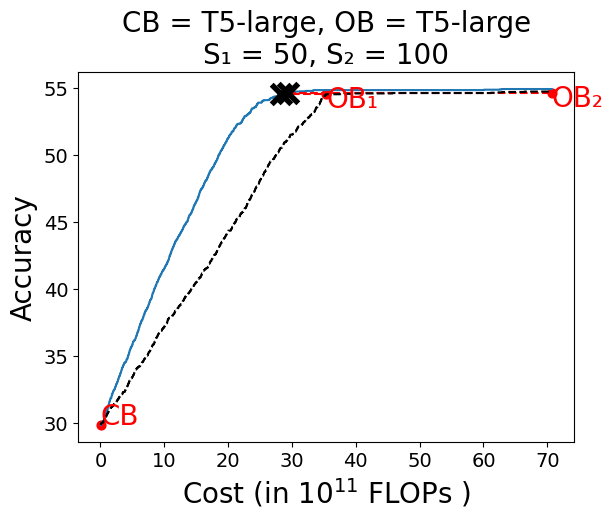}
    \end{subfigure}
    
    \caption{
    Accuracy-cost curves of the proposed cascading system (in blue) and baseline system (in black) for K=2 setting on NQ. 
    Red points correspond to the accuracy and cost values of the individual models $CB$, $OB_1$ (leveraging $S_1$ knowledge statements), and $OB_2$ (leveraging $S_2$ knowledge statements). 
    Points of intersection of red dashed lines drawn from $OB_1$ and $OB_2$ on the blue curve correspond to costs at which the cascading system achieves the same accuracy as $OB_1$ and $OB_2$ respectively. $OB_1$ and $OB_2$ are the same models but differ only in the $S$ value.
    }
    \label{fig:supp_acc_cost_curves_KI_2_large}    
\end{figure*}

In this setting, we first infer using the CB model and then conditionally use two knowledge iterations with the open-book model. 
In other words, if the CB model is not sufficiently confident then we use the OB model with $S_1$ knowledge statements ($OB_1$), and if $OB_1$ is not sufficiently confident then we use the OB model with $S_2$ knowledge statements ($OB_2$).
Figure \ref{fig:supp_acc_cost_curves_KI_2_base} and Figure \ref{fig:supp_acc_cost_curves_KI_2_large} show accuracy-cost curves of different configurations for this setting.

In this setting, our method achieves much larger efficiency improvements than the K=1 setting. 
For example, in case of (CB=T5-large, OB=T5-base, $S_1$=1, $S_2$=2), $OB_2$ achieves $37.48\%$ accuracy at the cost of $0.4 \times 10^{11}$ FLOPs and cascading system achieves the same accuracy at just $0.13 \times 10^{11}$ FLOPs.
Moreover, we achieve efficiency improvements over $OB_1$ model also; in the same case, $OB_1$ achieves $32.88\%$ accuracy at $0.2 \times 10^{11}$ FLOPs and cascading system achieves the same accuracy at just $0.09 \times 10^{11}$ FLOPs.

In case of (CB=T5-large, OB=T5-base, $S_1$=10, $S_2$=20), cascading system achieves the same accuracy as $OB_2$ at just $39.39\%$ of $OB_2$'s computation cost. Specifically, $OB_2$ achieves $48.61\%$ accuracy at the cost of $4.04 \times 10^{11}$ FLOPs and cascading system achieves the same accuracy at just $1.59 \times 10^{11}$ FLOPs.
Our method achieves efficiency improvements over $OB_1$ model also; in the above case, $OB_1$ achieves $46.78\%$ accuracy at the cost of $2.019 \times 10^{11}$ FLOPs and cascading system achieves the same accuracy at just $1.17 \times 10^{11}$ FLOPs.
Similarly, in case of (CB=T5-large, OB=T5-base, $S_1$=20, $S_2$=50), cascading system achieves the same accuracy as $OB_2$ at just $33.97\%$ of $OB_2$'s cost.

\section{Re-using encoded representations of previous knowledge iterations}

Fusion-in-Decoder model \cite{izacard-grave-2021-leveraging} is one of the top performing open-book models.
It first computes the representation of question + knowledge for each knowledge statement (with a fixed number of tokens) independently using the encoder and then concatenates these representations and passes it to the decoder for making the prediction.
As mentioned before, the encoded representations of knowledge of $(k-1)^{th}$ iteration can be stored and re-used in $k^{th}$ knowledge iteration. 
This implies that only the new knowledge needs to be encoded in the next iteration.
This will further improve the computational efficiency. However, it requires auxiliary space for storage. We will explore this trade-off between auxiliary space (for storing encodings of knowledge) and inference cost as a future work.

\section{Related Work on Efficiency in NLP}
With the introduction of large-scale pre-trained language models, the efficiency topic has attracted a lot of research attention. 
Efficiency is being studied from diverse lenses such as training data efficiency \cite{lewis-etal-2019-unsupervised,schick-schutze-2021-exploiting, varshney-etal-2022-unsupervised, Wang2021TowardsZL,ben-zaken-etal-2022-bitfit},
evaluation efficiency \cite{rodriguez-etal-2021-evaluation, varshney-etal-2022-ildae}, 
parameter tuning efficiency \cite{li-liang-2021-prefix,pmlr-v97-houlsby19a}, 
retrieval efficiency \cite{zhao-etal-2021-sparta,luo2022study}, on-disk memory efficiency \cite{pmlr-v133-min21a, izacard2020memory}, and 
inference efficiency.
In this work, we focus on inference efficiency of ODQA reader and propose an approach that utilizes both the `closed-book' (relying on the knowledge already present in the model parameters) and the `open-book' inference (leveraging external knowledge).
Furthermore, instead of using a large fixed number of passages for open-book inference, we dynamically read the external knowledge in multiple `knowledge iterations'.
Our method is also related to selective prediction \cite{kamath-etal-2020-selective, varshney-etal-2022-investigating, varshney-etal-2022-towards,garg-moschitti-2021-will,xin-etal-2021-art} as the system first tries to answer using the low-cost inference and then selectively utilizes the high-cost inference only when it is not sufficiently confident in its prediction.
Thus, by carefully deciding when external knowledge is required and whether the current amount of external knowledge is sufficient to answer a question correctly, the computational efficiency of the reader system can be considerably improved while maintaining the high prediction accuracy.

\end{document}